%% file: a-contrario_clustering.tex
\documentclass[11pt]{article}
\usepackage{caption}
\usepackage{subcaption}
\usepackage{amsfonts}
\usepackage{bbm}
\usepackage{amsmath}
\usepackage[ruled,vlined,linesnumbered]{algorithm2e}
\usepackage[section]{placeins}
\usepackage{epsfig}
\usepackage{amssymb}
\usepackage[numbers]{natbib}
\usepackage{graphicx}
\usepackage{amsthm}
\usepackage{booktabs}
\usepackage{etoolbox}
\usepackage{adjustbox}
\usepackage{multirow}
\usepackage[round-mode=places,detect-weight=true,detect-inline-weight=math]{siunitx}

\theoremstyle{definition}
\newtheorem{definition}{Definition}[section]

\SetKwInput{KwInput}{Input}                
\SetKwInput{KwOutput}{Output}              

\begin{document}

\title{Seed-Guided Semi-Supervised Clustering by A-Contrario Anomaly Detection}

\author{Nassir Mohammad\thanks{Email: nassir.mohammad@airbus.com} \thanks{Project code is available at https://github.com/M-Nassir/clustering} \\ 
\emph{Cyber Innovation Lab, Airbus, Newport, UK}
}
\maketitle

\begin{abstract}
This paper introduces a semi-supervised clustering framework grounded in the statistical duality between grouping principles and anomaly detection. We address the challenge of robust cluster definition in noisy environments---a task where partitioning algorithms often over-assign outliers and density-based methods remain sensitive to heuristic global parameters. Drawing on \textit{a-contrario} statistical reasoning and Gestalt proximity principles, we define a cluster as a maximal subset of data points containing no anomalies relative to a null hypothesis of uniform randomness. Central to this approach is the Perception algorithm, which utilises a principled expectation-based threshold ($\mathbb{E} < 1$) to identify outliers without manual parameter tuning. By treating clustering as the dual of anomaly detection, we employ an iterative ``clustering-by-exclusion'' mechanism. The algorithm is seed-guided, leveraging minimal user-provided labels to initialise robust cluster medians and form initial groups, which are subsequently expanded by admitting non-anomalous points. This approach naturally isolates fringe points, isolated noise, and emerging unknown clusters. We evaluate the method on synthetic and real-world benchmarks, including image and text datasets represented through raw, linear-reduced, and neighbourhood-preserving embeddings. Results demonstrate that with as few as 10--30 seeds per cluster, the proposed method achieves competitive and often very strong performance under a practical low-tuning benchmarking protocol, while maintaining linear scalability with respect to both observations and dimensionality for a fixed number of seeded clusters and iterations.
\end{abstract}

\FloatBarrier
\section{Introduction} 
\label{section:introduction} 

Grouping is a fundamental cognitive process, particularly critical in the interpretation of visual information. This is evidenced by Gestalt Psychology, where local groupings (partial gestalts) repeatedly fuse to culminate in global perceptions \cite{Morel07}. Consequently, cluster analysis has become a cornerstone of modern data mining, with applications ranging from malware detection in cybersecurity and topic modelling in document processing to gene expression analysis in molecular biology.

Despite its importance, clustering remains an inherently ill-posed problem. Conventional definitions often describe clusters as regions where internal points are \textit{similar} and external points are \textit{dissimilar}, yet this fails to provide a computable criterion for where a group begins and ends. A significant challenge in current methodologies is the ``forceful lumping'' of disparate observations into existing groups, which often masks underlying noise or emerging anomalous structures. This is particularly problematic in partitioning algorithms that lack a native mechanism for outlier rejection.

Furthermore, many established algorithms rely on heuristic parameters, such as the number of clusters $k$ or density thresholds $\varepsilon$, which are difficult to infer without prior knowledge of the data's scale. In practice, such parameters specify the resolution at which the clustering process is carried out. Practitioners often estimate these values through iterative trial-and-error, effectively transforming unsupervised tasks into informal semi-supervised processes. Recent shifts towards constrained clustering have sought to formalise this guidance using labelled samples or instance-level relationships \cite{WagstaffCRS01}. However, these methods frequently inherit the underlying sensitivities of their parent algorithms to outliers and rigid geometric assumptions, without necessarily improving performance.

Recent work has also moved beyond a strict separation between partitioning and density-based clustering. Hybrid density-partitioning and subspace-density methods combine density estimates, density peaks, feature or subspace weighting, and partitioning objectives in order to mitigate limitations of both global density thresholds and rigid centroid-based assignments. The motivation of the present work should therefore not be read as a claim that modern clustering remains confined to a simple partitioning-versus-density dichotomy. Rather, the gap addressed here is more specific: even when hybrid methods improve the discovery of density-aware structures, they do not necessarily provide an explicit statistical criterion for deciding whether a candidate point should be admitted into a seed-defined target cluster or rejected as inconsistent with all target clusters. Our contribution is therefore complementary: we introduce a seed-guided, anomaly-aware membership principle in which cluster expansion is governed by an \textit{a-contrario} admission/rejection test rather than by complete assignment to the nearest, densest, or globally optimised partition.

In the present work, we integrate the objectivity of automated discovery with the targeted intent of semi-supervised guidance. We introduce a formalised framework grounded in the duality between grouping principles and anomaly detection. At the computational level, we view clustering as the complement of anomaly detection under a chosen statistical model of meaningful structure. In the present paper, this broader principle is instantiated through an \textit{a-contrario} null hypothesis of uniform randomness together with a concrete iterative ``clustering-by-exclusion'' mechanism. In this paradigm, user-provided seeds initialise the process and define the target resolution, while the statistical model autonomously determines cluster membership and rejects fringe points.

The primary contributions of this paper are as follows:
\begin{itemize}
    \item \textbf{Statistical Duality Framework:} We formalise clustering as the statistical dual of anomaly detection at the computational level, and provide a computable instantiation of this principle grounded in the \textit{a-contrario} framework.
    \item \textbf{Robust Perception Kernel:} We provide a reproducible implementation of an autonomous, self-thresholding anomaly detector that eliminates the need for manual parameter tuning.
    \item \textbf{Resilient Expansion Algorithm:} We introduce a seed-guided expansion process that effectively integrates expert intent while remaining robust to noisy initialisation, as inconsistent seeds are statistically ejected during the iterative refinement phase.
    \item \textbf{High-Dimensional Evaluation and Scalability:} We demonstrate that the method achieves competitive performance on synthetic and real-world benchmarks, including image and text datasets evaluated in raw, linear-reduced, and neighbourhood-preserving representations, while maintaining linear computational complexity with respect to both observations and dimensionality for fixed seeded-cluster and iteration counts.
\end{itemize}

The remainder of this paper is organised as follows: Section~\ref{section:relatedWork} reviews established paradigms and their semi-supervised variants. Section~\ref{section:computational_theory} details the computational theory and mathematical formulation. Section~\ref{section:clustering_algorithm} describes the algorithmic implementation. Finally, Section~\ref{section:example_results} presents experimental results and Section~\ref{section:conclusion} discusses limitations and future work.

\section{Related Work}
\label{section:relatedWork}

Clustering is a fundamental task in data mining, with established paradigms 
including hierarchical, density‑based, probabilistic, and centroid‑based 
methods. Despite the proliferation of new techniques, practitioners often 
favour $k$-means, DBSCAN, and Gaussian Mixture Models (GMMs) for their 
simplicity and widespread availability. However, systemic limitations persist, 
particularly regarding hyperparameter sensitivity, the ``curse of 
dimensionality'', and the difficulty of defining robust cluster boundaries in 
the presence of noise.

\subsection{Unsupervised Baselines}
Hierarchical methods \cite{Ward63} provide multi‑resolution insights but 
suffer from $O(n^2 \log n)$ time complexity and irreversible merge decisions, 
making them sensitive to initial noise. DBSCAN \cite{ester1996density} 
identifies arbitrary shapes and isolates noise; however, its performance often 
degrades in high‑dimensional spaces and it is highly sensitive to the 
$\varepsilon$ and $minPts$ parameters, which assume globally uniform density. 
Probabilistic methods such as GMMs offer flexible ellipsoidal modelling via 
Expectation‑Maximisation but are sensitive to initialisation and assume 
clusters follow a specific distribution. The $k$-means algorithm 
\cite{macqueen1967} remains the benchmark for efficiency but lacks a native 
mechanism for outlier rejection, often leading to distorted centroids as 
anomalies are forcefully absorbed.

\subsection{Hybrid Density–Partitioning and Subspace Density Methods}
Recent work has moved beyond a strict separation between partitioning and 
density‑based clustering. Hybrid density‑partitioning, projection, and subspace 
clustering approaches combine density estimates, density peaks, feature or 
subspace weighting, and partitioning objectives to mitigate limitations of both 
global density thresholds and rigid centroid‑based assignments. Representative 
examples include SDENK, an unbiased subspace density‑$k$ clustering method 
\cite{Zou2025SDENK}, weighted‑density subspace clustering methods such as 
``Weighted Density for The Win: Accurate Subspace Density Clustering'' 
\cite{10890745}, and DipNSub, which uses dip-test-based projections to identify 
subspace cluster structure \cite{Bauer2023DipNSub}. These methods primarily aim 
at structural discovery: identifying density‑relevant subspaces, density peaks, 
or partitions that better reflect the intrinsic organisation of the data. In 
contrast, the present work assumes seed-defined target groups and focuses on 
the statistical admission or rejection of candidate observations relative to 
those groups.

\subsection{Semi‑Supervised and Constrained Clustering}
Semi‑supervised clustering incorporates prior knowledge to align algorithmic 
outcomes with user intent. Constrained clustering commonly uses instance‑level 
information, such as must‑link and cannot‑link pairs, as in Seeded $k$-means 
\cite{BasuBM02}, COP‑KMeans \cite{WagstaffCRS01}, and PCK‑Means 
\cite{basu_pckmeans_2004}. Semi‑supervised extensions of DBSCAN 
\cite{SSDBSCANLelisS09} and Gaussian Mixture Models \cite{ShentalBHW03} 
similarly incorporate labels or equivalence constraints to guide density or 
model estimation. Graph-based label-propagation methods, including label propagation and LabelSpreading \cite{ZhuGhahramani2002,ZhouBLWS2004}, provide another important semi-supervised paradigm: labels are diffused from labelled examples across a similarity graph or kernel. These methods are effective when the graph captures the class geometry, but they remain sensitive to graph or kernel construction and typically propagate labels to all observations rather than preserving an explicit rejected or unassigned set.

More recent approaches have explored stronger structural assumptions about data 
geometry and density. Examples include density‑based semi‑supervised 
hierarchical clustering via low‑density separation (DBSSHC) 
\cite{JiangQFG2026DBSSHC}, semi‑supervised clustering under a compact‑cluster 
assumption (CSSC) \cite{JiangZMD2023CSSC}, and unsupervised and 
semi‑supervised clustering via density‑ and distance‑based label propagation 
and assignment (DDBC) \cite{Jiang2025DDBC}. These methods represent 
complementary paradigms: density‑guided hierarchical splitting, 
compact‑cluster modelling, and density/distance‑based propagation. Compared 
with classical constrained and graph-propagation methods, they improve
flexibility, but they generally still depend on the quality of the learned representation and on neighbourhood, density, graph, or propagation parameters.

Further semi‑supervised paradigms extend clustering into structured or fuzzy 
representational frameworks. Semi‑supervised spectral clustering 
\cite{KamvarKM03} incorporates pairwise constraints into the graph Laplacian 
to guide label propagation over a similarity graph, but inherits sensitivity 
to graph construction and spectral decomposition parameters. Semi‑supervised 
subspace clustering \cite{LiV15_subspace} combines sparse or low‑rank 
subspace representations with supervision constraints to recover 
union‑of‑subspace structure, but requires the subspace dimension or sparsity 
level to be specified. Semi‑supervised fuzzy clustering \cite{PedryczSFCM} 
relaxes hard cluster assignments by allowing graded memberships guided by 
labelled examples, but introduces membership degree parameters and relies on a 
fixed partition count. 

The proposed framework uses supervision differently from these paradigms. Seed
labels define the target groups, but they are not propagated to all observations
and they are not used to construct a complete partition. Instead, each candidate
observation is tested for statistical admissibility relative to the
seed-defined cluster centre using the fixed \textit{a-contrario} criterion.
Observations that fail this test for every seeded cluster remain rejected or
unassigned. In this sense, the method does not require a pre-specified subspace
dimension, a fuzzy membership degree, or a graph construction step; it provides
a complementary admission-and-rejection layer that can protect seeded clusters
from fringe observations, noise, or unseeded structures. Hybrid
density-partitioning or subspace-density methods may still supply useful
representations or candidate structures, while the proposed mechanism evaluates
whether individual observations are statistically coherent members of the
seeded target groups. For a comprehensive taxonomy and recent advances in
semi-supervised constrained clustering, we refer the reader to the survey by
González-Almagro et al. \cite{GonzalezAlmagroPPCG25}.

\subsection{Deep Clustering}
Deep clustering methods use neural networks to learn latent representations
that are optimised jointly with clustering objectives, as reviewed in recent
surveys \cite{DeepClustering2025}. Representative examples include Deep
Embedded Clustering (DEC) \cite{XieGF16_DEC}, Deep Clustering Networks (DCN)
\cite{Yang2017_DCN}, and DipEncoder \cite{Leiber2022DipEncoder}, which
encourages multimodal latent representations. These approaches are especially
relevant for high-dimensional data because clustering quality often depends as
much on the representation as on the final grouping rule. However, many deep
clustering models remain partition-oriented: they assign observations to latent
clusters without providing an explicit statistical test for whether a candidate
observation should be admitted to, or rejected from, a target cluster.

\subsection{Outlier-Aware Clustering}
Outlier-aware clustering methods address a related but distinct limitation of
complete partitioning. Methods such as $k$-means-- \cite{ChawlaGionis2013} and
COR \cite{LiuCOR2018} jointly consider clustering and outlier removal, but
typically introduce additional non-trivial choices, such as the number or
treatment of observations to be removed. More recent work has considered
heterogeneous anomaly structure, including scattered outliers, clustered
anomalous groups, and multi-granular outlier patterns. For example, MGOD
studies multi-granular outlier detection with clustlier analysis
\cite{MGOD2024}, while hierarchical reference-set approaches distinguish
scattered and clustered outliers \cite{HierRefSets2026}. These methods address
a related but distinct problem: they aim to detect or characterise different
forms of anomalous structure, whereas the present paper uses anomaly detection
as a membership criterion for seeded clustering. This distinction is relevant
because the rejected set in anomaly-aware clustering may itself contain
heterogeneous structure, including isolated noise, coherent unseeded groups, or
multi-granular clustliers.

\section{A Computational Theory for Clustering}
\label{section:computational_theory}

The goal of the present work is to draw insights from the human visual process of object grouping to develop a robust algorithmic framework for large-scale data analysis. We treat clustering as an information-processing task structured according to Marr's tri-level hypothesis \cite{Mar82}. This top-down hierarchy disciplines the analysis to delineate the statistical logic of the task from its specific algorithmic implementation, ensuring the method is both theoretically grounded and computationally efficient.

\subsection{The Tri-level Framework}
\label{subsection:tri-level}

Marr's analysis posits that a complex information-processing system cannot be understood through a single level of explanation; instead, it requires three distinct levels of formalisation that separate the logic of the problem from the mechanics of the solution:

\begin{enumerate}
    \item \textbf{The Computational Level:} This level deals with the logic of the task: \textit{what} information is extracted and \textit{why} the computation is appropriate. For clustering, we introduce the \textit{a-contrario} principle as a fundamental constraint grounded in the Helmholtz principle of human perception \cite{Morel07}. This principle posits that a pattern is meaningful only when its occurrence is statistically unexpected under a null hypothesis of uniform randomness. This logic is appropriate because it provides an objective, computable boundary for clusters based on statistical significance, effectively differentiating true structural groupings from stochastic fluctuations without the need for heuristic parameters such as a pre-defined number of clusters $k$ or global density thresholds.
    
    \item \textbf{The Algorithmic Level:} This level is concerned with \textit{how} the computations are carried out. We utilise the Euclidean distance to the cluster median as our primary representation, mapping high-dimensional data into a robust one-dimensional distance space. The transformation is achieved through an iterative ``clustering-by-exclusion'' process that grows clusters from initial seeds by admitting only those points that maintain the statistical integrity of the group. The procedure is designed to remain computationally scalable for fixed seeded-cluster and iteration counts, with the full complexity analysis provided in Section~\ref{subsection:complexity}.
    
    \item \textbf{The Implementation Level:} This level focuses on the physical realisation of the algorithm. We provide an implementation in Python that preserves the theoretical efficiency of the algorithmic design. By utilising vectorised operations and efficient memory management, the implementation ensures that the linear complexity translates into practical scalability. This is suitable for high-dimensional datasets and large-scale industrial applications where computational overhead is a critical constraint, regardless of the underlying hardware architecture.
\end{enumerate}

This structured approach allows us to reconcile automated discovery with semi-supervised interaction. By permitting a small fraction of labelled seeds, we provide the algorithm with the necessary \textit{resolution} (scale of interest) and \textit{intent} (specific groups of interest) to guide the iterative process towards user-defined objectives, while the \textit{a-contrario} logic handles the objective membership boundaries.

\subsection{Theoretical Formulation of Clustering Duality}
\label{subsection:clusteringTheory}

The clustering problem is inherently ill-posed; without principled constraints, any partition of data remains arbitrary. Building upon the perceptual logic established in Section~\ref{subsection:tri-level}, we formalise the relationship between clustering and anomaly detection through the following definitions. The duality is understood here at the computational level: clustering is treated as the recovery of statistically coherent sets by excluding observations that are anomalous under a chosen null model. The specific median-based representation, uniform null hypothesis, discrete counting model, and overlap handling adopted below constitute one concrete algorithmic instantiation of this broader principle.

\begin{definition}[Anomaly]
Let $\mathcal{H}_0$ be a null hypothesis representing a uniform random distribution of data. An observation $x$ is defined as an anomaly if its expectation of occurrence satisfies $\mathbb{E}(x \mid \mathcal{H}_0) < 1$. This signifies that the occurrence is statistically unexpected, as such a configuration would appear less than once on average under the null hypothesis of pure randomness.
\end{definition}

\begin{definition}[Cluster]
A cluster $C \subset \mathcal{X}$ is a maximal subset of data points such that every constituent point $x \in C$ is non-anomalous with respect to the group's internal distribution. Formally, $C$ must satisfy the condition:
\begin{equation}
    C = \{x \in \mathcal{X} \mid \mathbb{E}(x \mid C) \ge 1\}
\end{equation}
subject to the maximality constraint: there exists no $x' \in \mathcal{X} \setminus C$ such that the augmented set $\{C \cup x'\}$ remains non-anomalous.
\end{definition}

This clustering-by-exclusion framework ensures that points are grouped only while they maintain statistical coherence. A significant advantage of this approach is that it is inherently self-scaling; because the expectation $\mathbb{E}$ is calculated relative to the internal distribution of $C$, the rejection threshold adapts to the local density of each group. In dense regions, the expectation for a point to exist at a relatively large distance is low, resulting in a tighter boundary. Conversely, in sparser regions, the model naturally accommodates a larger spread, provided the configuration remains more likely than a random occurrence.

By allowing the statistical model to dictate these boundaries, the framework avoids the ``forceful lumping'' characteristic of partitioning algorithms like $k$-means, which lack a native mechanism to exclude outliers. Furthermore, the maximality constraint prevents arbitrary over-fragmentation by requiring the cluster to grow until it reaches a statistically significant boundary. In multi-cluster environments, clusters compete for observations based on their relative statistical coherence. For fixed seeds and a fixed deterministic cluster ordering, the resulting assignment procedure is reproducible. However, we do not claim a unique global partition independent of the initial supervision or overlap-resolution policy; rather, the method converges empirically to a stable seeded partition in the experiments considered here.

Consequently, disparate or unknown observations are naturally left unassigned. This facilitates iterative discovery, as these ``excluded'' points can be subsequently evaluated as potential seeds for new clusters, ensuring that noise does not distort the representation of established groups.

\subsection{Representation and Seed-Guidance}
\label{section:clustering_representation}

The choice of representation determines which Gestalt grouping principles are highlighted and how the statistical expectation is calculated. We utilise the Euclidean distance to the cluster median as our primary representation. The median is selected for its robust properties, possessing a 50\% breakdown point compared to the high sensitivity of the mean to outliers. This ensures that the cluster centre remains stable during the iterative expansion, even if the initial seed set contains significant noise.

Crucially, this representation maps high-dimensional observations into a one-dimensional distance space relative to the cluster centre. This dimensionality reduction allows the \textit{a-contrario} framework to operate on the distribution of distances, reducing the sparsity issues typically associated with the ``curse of dimensionality'' in raw feature spaces. However, it does not eliminate the need for a meaningful metric representation. If pairwise or centre-relative distances are already degraded by high-dimensional concentration, sparsity, or an unsuitable metric, the expectation test may become unstable or uninformative. We therefore do not claim a universal maximum dimensionality bound for the Gestalt-based criterion; its stability depends on the intrinsic dimension, representation geometry, noise level, and distance metric. At the same time, the centre-distance representation introduces a bias toward compact, centre-based clusters. As a consequence, the present formulation is less naturally suited to strongly non-convex, multi-modal, or manifold-shaped classes unless a more suitable representation is used upstream. This bias is nevertheless useful in many practical settings. As in centroid-based methods such as $k$-means, a centre-relative representation can provide an effective summary when target groups are compact, or when an upstream embedding makes them approximately compact. 

Acknowledging that clustering is an inherently interactive process, we utilise seed-guidance to resolve the ambiguity of cluster resolution. A small fraction of user-provided labels $\mathcal{X}_L \subset \mathcal{X}$ provides the algorithm with three critical components:
\begin{enumerate}
    \item \textbf{Intent}: Directing the algorithm toward specific groups of interest amongst a vast number of potential statistical partitions.
    \item \textbf{Initialisation}: Defining the initial cluster medians to anchor the first iteration of the expectation-based membership test.
    \item \textbf{Noise Tolerance}: The iterative process naturally accounts for erroneous supervision; mislabelled or inconsistent seeds are statistically ejected if they satisfy the anomaly condition ($\mathbb{E} < 1$) during the refinement phase.
\end{enumerate}
Accordingly, the framework should be understood as replacing global hyperparameter search with limited, interpretable supervision rather than eliminating user input altogether.

Observations rejected by all seeded clusters are retained as an explicit
anomaly or unassigned set. This separates the seeded target structures from
points that are statistically inconsistent with them, and leaves such points
available for subsequent inspection or interactive refinement. The rejected set
is not automatically decomposed by the present method: it may contain isolated
noise, coherent unseeded groups, or multi-granular ``clustlier'' structures. An
unseeded dense group is therefore rejected only insofar as it is inconsistent
with every seeded cluster under the chosen representation; if such a group is
seeded, or is statistically coherent with a seeded cluster, it may instead be
admitted.

\section{Algorithmic Realisation}
\label{section:clustering_algorithm}

The proposed algorithm transforms a dataset into a set of coherent clusters by iteratively refining memberships through statistical exclusion. This approach prioritises computational efficiency, ensuring the method scales linearly with both the number of observations and the feature dimensionality.

\subsection{The Perception Anomaly Detector}
The core of the clustering method is the \textit{Perception} anomaly detection algorithm \cite{nassir2021anomaly}. It operates on the principle that an observation is anomalous if its occurrence is unexpected under a null hypothesis $\mathcal{H}_0$ of uniform randomness. 

To formalise this, continuous distance measurements are transformed into a discrete representation suitable for combinatorial analysis. This is achieved by shifting the decimal point to the right---up to a maximum precision $\alpha$---until the required numerical resolution is captured in an integer space. Specifically, the distance $D$ is mapped to a discrete space $V$ by shifting the decimal point to eliminate fractional components. In the Perception implementation used throughout this paper, $\alpha$ acts as a cap on the retained decimal precision rather than a fixed effective precision in every case. Thus, if the observed distances are already integers, no decimal shift is needed, whereas if the finest observed precision is two decimal places, retaining two decimal places is sufficient. Because the discrete \textit{a-contrario} counting model operates on integerised distances, some finite decimal cut-off must be chosen when mapping continuous values into discrete space. We therefore treat $\alpha$ as an implementation-level numerical precision setting required to \textit{integerise} the data, rather than as a heuristic tuning parameter or decision threshold. The default value $\alpha=4$ was used as a practical upper bound that retained sufficient significant digits for the distance representation in our experiments. This quantisation allows the algorithm to treat distances as discrete unit-intervals from the cluster centre, thereby enabling the use of combinatorial probability to evaluate statistical significance.

Consider a set of such observations represented as discrete non-negative integers $V=[v_1, v_2, \dots, v_W]$, where $W$ is the number of discrete bins and $S = \sum V$ is the total sum of indicators (total events). Under $\mathcal{H}_0$, the expected number of $n$-tuples in a given configuration, $\mathbb{E}(C_n)$, is calculated as:
\begin{equation}
\label{equation:anomaly_detector}
\mathbb{E}(C_n) = \binom{S}{n} \frac{1}{W^{n-1}}
\end{equation}
An event is perceived as an anomaly if $\mathbb{E}(C_n) < 1$. This threshold is both principled and data-driven; if an event is unexpected under the random model but appears in the data, it is, by definition, anomalous. This mechanism allows the algorithm to adapt to the specific density and scale of each cluster without manual parameter tuning. The implementation details for the training and testing phases are provided in Algorithms \ref{algorithm:PerceptionTrain} and \ref{algorithm:PerceptionTest}.

\begin{algorithm}[H]
    \SetAlgoLined
    \caption{Perception Training Phase}\label{algorithm:PerceptionTrain}
    \DontPrintSemicolon
    \KwInput{Data matrix $X \in \mathbb{R}^{n \times d}$, distance metric $d(\cdot)$, precision $\alpha$}
    \KwOutput{Model parameters $\theta = \{S, W, \tilde{x}, \mu, \sigma, \text{med}_V\}$}
    
    \BlankLine
    1. Compute robust scaling parameters $\mu, \sigma$ and standardise $X \to X'$\;
    2. Compute marginal median $\tilde{x}$ where $\tilde{x}_j = \text{median}(\{X'_{i,j}\}_{i=1}^n)$ for $j=1 \dots d$\;
    3. Compute distance vector $D$ where $D_i = d(X'_i, \tilde{x})$ for $i=1 \dots n$\;
    4. Map to discrete space: $V = \lfloor D \cdot 10^\alpha \rfloor$\;
    5. Calculate central deviation relative to training median: $V_{rel} = |V - \text{median}(V)|$\;
    6. $S \leftarrow \sum V_{rel}$; \quad $W \leftarrow \text{length}(V_{rel})$; \quad $\text{med}_V \leftarrow \text{median}(V)$\;
    \Return $\{S, W, \tilde{x}, \mu, \sigma, \text{med}_V\}$\;
\end{algorithm}

\begin{algorithm}[H]
    \SetAlgoLined
    \caption{Perception Test Phase (Anomaly Scoring)}\label{algorithm:PerceptionTest}
    \DontPrintSemicolon
    \KwInput{Point $x \in \mathbb{R}^d$, Model parameters $\theta$}
    \KwOutput{Anomaly score $Z$, Binary label $Y$}
    
    \BlankLine
    1. Standardise $x$ using $\mu, \sigma$ and compute distance $D_x = d(x', \tilde{x})$\;
    2. Discretise relative to training median: $n = \lfloor | (D_x \cdot 10^\alpha) - \text{med}_V | \rfloor$\;
    3. Compute expectation: $\mathbb{E} = \binom{S}{n} \frac{1}{W^{n-1}}$\;
    4. $Z \leftarrow -\frac{1}{S} \ln(\mathbb{E})$ \tcp*{Log-transformed score}\;
    \lIf{$Z > 0$}{$Y \leftarrow 1$ (Anomaly)} 
    \lElse{$Y \leftarrow 0$ (Normal)}
    \Return $Z, Y$\;
\end{algorithm}

\subsection{Iterative Seed-Guided Expansion}
The clustering process utilises a small, user-provided subsample of labelled data, $\mathcal{X}_L$, to initialise the procedure. These seeds define the initial cluster medians, while all unlabelled points are initially classified as anomalies (label $-1$). The procedure seeks a stationary statistical equilibrium through two alternating phases: \textit{ejection} and \textit{admission}, as detailed in Algorithm \ref{algorithm:Clustering}.

During the \textit{ejection} phase, a Perception model is fitted to the current members of a cluster; any observation found to be anomalous relative to the group's evolving distribution ($\mathbb{E} < 1$) is removed. In the \textit{admission} phase, the model is re-fitted to the pruned group, and all currently unassigned points are tested for membership. To resolve overlaps under this greedy assignment scheme, clusters are processed in an order determined by their statistical coherence (Step 2). This is a heuristic ordering rule rather than a quantity derived directly from the underlying \textit{a-contrario} model. If an observation would satisfy the non-anomaly condition ($\mathbb{E} \ge 1$) for multiple groups, processing compact clusters first reduces the tendency of more diffuse groups to absorb ambiguous points prematurely. The empirical effect of this ordering rule is evaluated in Section~\ref{subsection:overlap_ablation}.

For fixed seeds, fixed cluster ordering, and deterministic overlap handling,
each iteration induces a deterministic update of the label vector $L$. The
practical implementation always terminates in finite time because the loop is
capped at $t_{\max}$, and a fixed point is declared when no labels change
between successive iterations. Accordingly, the method is deterministically
reproducible for a fixed seeded configuration, but it does not imply a unique
global solution across all possible seed sets. In all experiments,
$t_{\max}$ was fixed to a conservative value of 100 and used only as a safety
cap; termination was instead governed by the fixed-point criterion, which was
reached well before this limit in practice.

No initial proximity graph is constructed in the present implementation. Thus, the method does not require a neighbourhood size, $minPts$ value, graph connectivity threshold, or minimal structural constant analogous to those used in graph- or density-connectivity methods. The main implementation choices are the seed set and the metric/representation in which centre distances are computed. The decision threshold itself remains fixed at $\mathbb{E}<1$.

\begin{algorithm}[H]
    \SetAlgoLined
    \caption{Seed-Guided Semi-Supervised Clustering}\label{algorithm:Clustering}
    \SetKwInOut{Input}{Input}
    \SetKwInOut{Output}{Output}
    \KwInput{$X$ (Data), $X_L$ (Labelled seeds), $t_{max}$ (Max iterations)}
    \KwOutput{$L$ (Labels), $M$ (Membership scores)}
    \BlankLine
    1. Initialise $L$ for unlabelled points as $-1$; labels for $X_L$ as per user input\;
    2. Order clusters by minimum sum of squared error (compactness)\;
    \For{$counter = 1$ \KwTo $t_{max}$}{
        \ForEach{cluster $C_k$}{
            3. \textbf{Ejection Phase}:\;
            Train Perception model $P_k$ on current members of $C_k$ (Alg. \ref{algorithm:PerceptionTrain})\;
            Identify and remove points $x \in C_k$ where $Z(x \mid P_k) > 0$\;
            Set $L(x) \leftarrow -1$ for all ejected points\;
            
            \BlankLine
            4. \textbf{Admission Phase}:\;
            \textit{Re-train} $P_k$ using the pruned set $C_k$ to update $S, W, \text{and } \tilde{x}$\;
            \ForEach{unassigned point $x_a$ where $L(x_a) = -1$}{
                Test $x_a$ using Algorithm \ref{algorithm:PerceptionTest} against updated $P_k$\;
                \If{$Y(x_a \mid P_k) = 0$}{
                    Assign $x_a$ to $C_k$ and set $L(x_a) \leftarrow k$\;
                }
            }
        }
        \If{no label changes occurred in $L$}{Break\;}
    }
    5. Compute final membership scores $M$ based on log-expectation\;
    \Return $L, M$\;
\end{algorithm}

\subsection{Computational Complexity}
\label{subsection:complexity}

The computational efficiency of the proposed framework is a key advantage for large-scale applications. Let $N$ denote the number of observations, $D$ the dimensionality of the representation used by the clustering algorithm, $K$ the number of seeded target clusters, and $t$ the number of admission/ejection iterations, with $t \leq t_{\max}$. Training a Perception model for one cluster is linear in the number of points assigned to that cluster and in $D$, because it requires robust scaling, marginal median estimation, and centre-distance computation. Testing candidate points for admission is likewise linear in the number of tested observations and in $D$.

In the worst case, each iteration may test all observations against each seeded cluster, giving a time complexity of $O(tKND)$. When $K$ and $t$ are small relative to $N$, as in the reported experiments, this behaves linearly in both the number of observations and the feature dimensionality. The space complexity is $O(ND + KN)$ when storing the data representation, labels, membership scores, and per-cluster temporary score vectors. The method does not store a full pairwise distance matrix or neighbourhood graph, avoiding the $O(N^2)$ memory cost common to graph-based clustering methods. Any dimensionality-reduction preprocessing, such as PCA, TruncatedSVD, or UMAP, is external to the clustering algorithm and its computational cost should be accounted for separately.

\section{Experimental Results}
\label{section:example_results}

This section evaluates the proposed algorithm across synthetic and real-world
benchmark datasets. We compare against established and recent reproducible
baselines from several method families. The benchmark was designed to prioritise
reproducibility and practical comparability: included baselines have stable,
directly callable implementations and are evaluated under a common low-tuning
protocol. Abbreviations used in the result tables are given in parentheses, and
Ours denotes the proposed method:
\begin{enumerate}
    \item \textit{Unsupervised}: $k$-means (KM), DBSCAN, Agglomerative Clustering (Agg), Gaussian Mixture Models (GMM), and DipNSub.
    \item \textit{Semi-supervised}: Seeded $k$-means (S-KM), Constrained $k$-means (C-KM), COP-KMeans (COP-KM), and LabelSpreading (LS).
    \item \textit{Deep Learning}: Deep Embedded Clustering (DEC) and DipEncoder (DipEnc).
\end{enumerate}

To ensure reasonable baseline performance, all methods requiring it were
provided with the ground-truth number of clusters ($k$). All baselines utilised
standard scikit-learn \cite{scikitlearn}, ClustPy
\cite{leiber2023benchmarking}, or established open-source implementations with
their default settings unless otherwise required by the evaluation protocol. We
intentionally maintained default hyperparameters for these baselines to reflect
the practical difficulty of manual parameter tuning in real-world scenarios,
where the underlying data distribution and noise levels are typically unknown.
Accordingly, the present study should be interpreted as a practical low-tuning
comparison with reproducible baselines rather than an exhaustive
hyperparameter-optimised benchmark.

\subsection{Evaluation Methodology and Robustness}
Evaluating clustering algorithms is inherently challenging, as many established metrics inadvertently favour specific optimisation criteria. Notably, total partitioning algorithms often appear to outperform anomaly-aware methods under conventional metrics. This discrepancy arises because many benchmark datasets enforce a global assignment, leaving no provision for anomalous or outlying points. Consequently, even the ground truth may fail to reflect the underlying statistical structure of the data. For instance, in synthetic Gaussian mixtures, observations in the extreme tails are reasonably considered outliers relative to the core density, yet they are rarely labelled as such in standard benchmarks (see Figure~\ref{fig:2d_gaussian_zoomed}).

Furthermore, a single dataset can often yield multiple valid partitions
depending on the feature-space resolution. An algorithm may discover
statistically significant groupings that are informative yet do not align with
predefined human labels. While expert human evaluation remains the gold standard
for resolving such ambiguity, its resource-intensive nature makes it
impractical for large-scale validation. Therefore, this work adopts a dual
evaluation strategy: controlled synthetic benchmarks are first used to
illustrate the statistical patterns and anomaly-rejection behaviour of the
algorithm. This is followed by quantitative evaluation using external
validation metrics to benchmark performance against reproducible unsupervised,
semi-supervised, subspace/projection, label-propagation, and deep clustering
baselines. Where applicable, reported quantitative scores are averaged over ten
independent runs with randomised seed selection.

\subsection{One-dimensional Synthetic Data}
\label{section:1d_results}

The first dataset, \textit{1d\_gauss}, evaluates the algorithm's ability to distinguish core cluster structures from heterogeneous noise. It consists of three Gaussian-distributed clusters of sizes $[\num{10000}, \num{5000}, \num{2500}]$ with centres at $[0, 50, 100]$ and standard deviations of $[1, 3, 8]$, respectively. The dataset is further complicated by the addition of isolated local anomalies, several deliberately mislabelled examples, and a small, globally anomalous cluster for which no seeds are provided. 

\begin{figure}[htbp]
    \centering
    \includegraphics[width=1\linewidth]{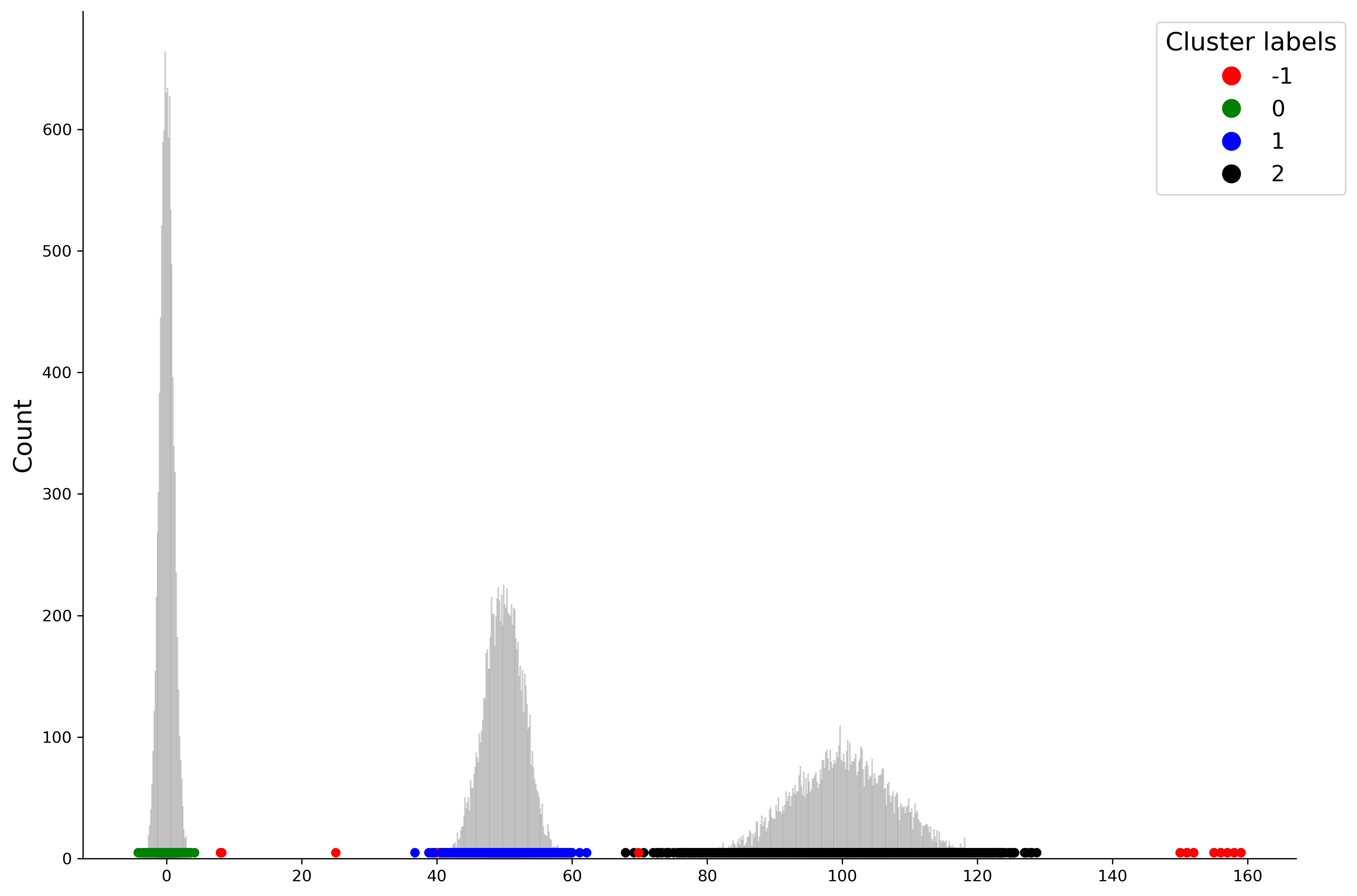}
    \caption{Ground-truth distribution of the \textit{1d\_gauss} dataset. The plot and accompanying histogram reveal three primary clusters, isolated anomalies, and a small, disparate anomalous group.}
    \label{fig:1d_gaussian_y_true}
\end{figure}

Figure~\ref{fig:1d_gaussian_y_true} visualises the ground-truth distribution. As established in our methodology, standard Gaussian distributions contain observations in their extreme tails that are statistically unexpected; however, traditional ground-truth labels rarely categorise these as outliers. By contrast, our \textit{a-contrario} framework explicitly identifies these ``fringe points'' as anomalies based on the Helmholtz Principle of unexpectedness.

To initialise the seed-guided process, a minimal fraction of the data---comprising only 35 examples (0.2\%)---was randomly selected as seeds, excluding the small anomalous cluster. These seeds represent the sparse domain knowledge typically available in interactive exploratory sessions. Notably, the algorithm does not require seeds for every existing group; unseeded clusters are simply reported as anomalies, allowing for subsequent discovery.

\begin{figure}[htbp]
    \centering
    \includegraphics[width=1\linewidth]{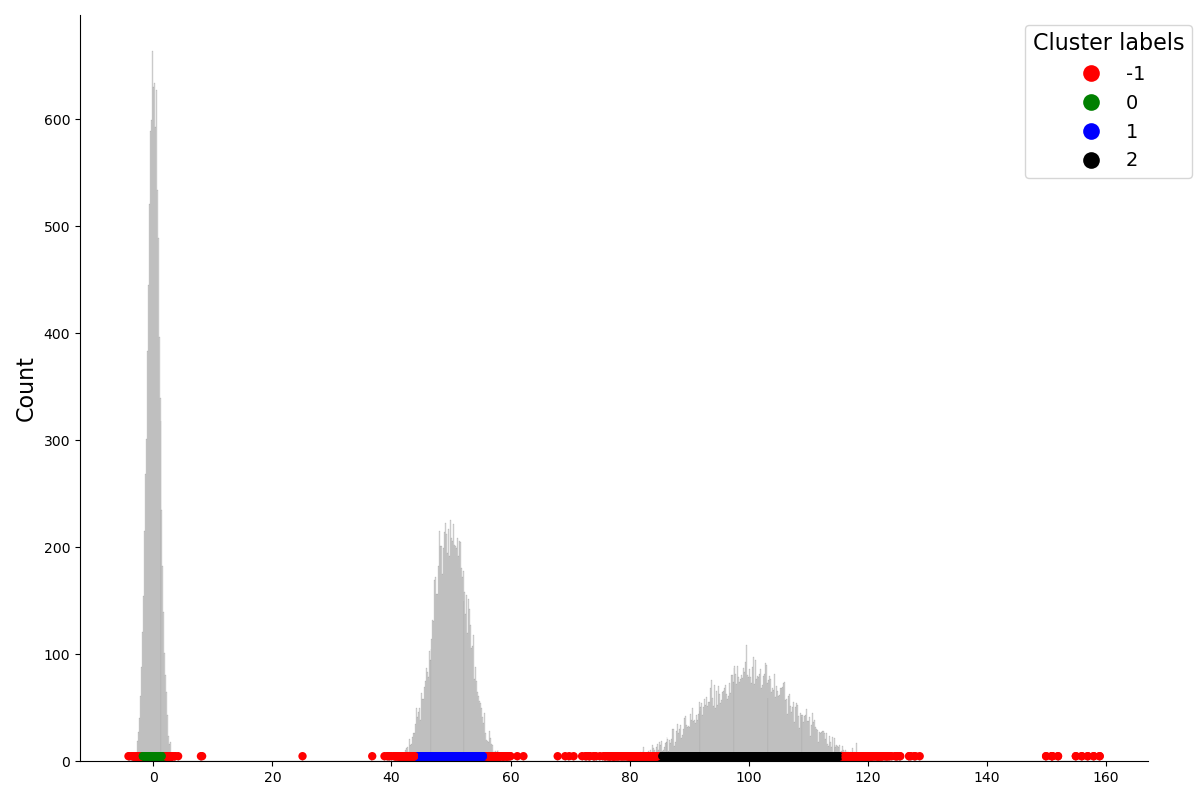}
    \caption{Clustering results of the proposed method. The algorithm identifies the primary clusters while relegating the unseeded anomalous group to the outlier category, preserving cluster purity.}
    \label{fig:1d_nassir_results}
\end{figure}

\begin{figure}[htbp]
    \centering
    \includegraphics[width=0.9\linewidth]{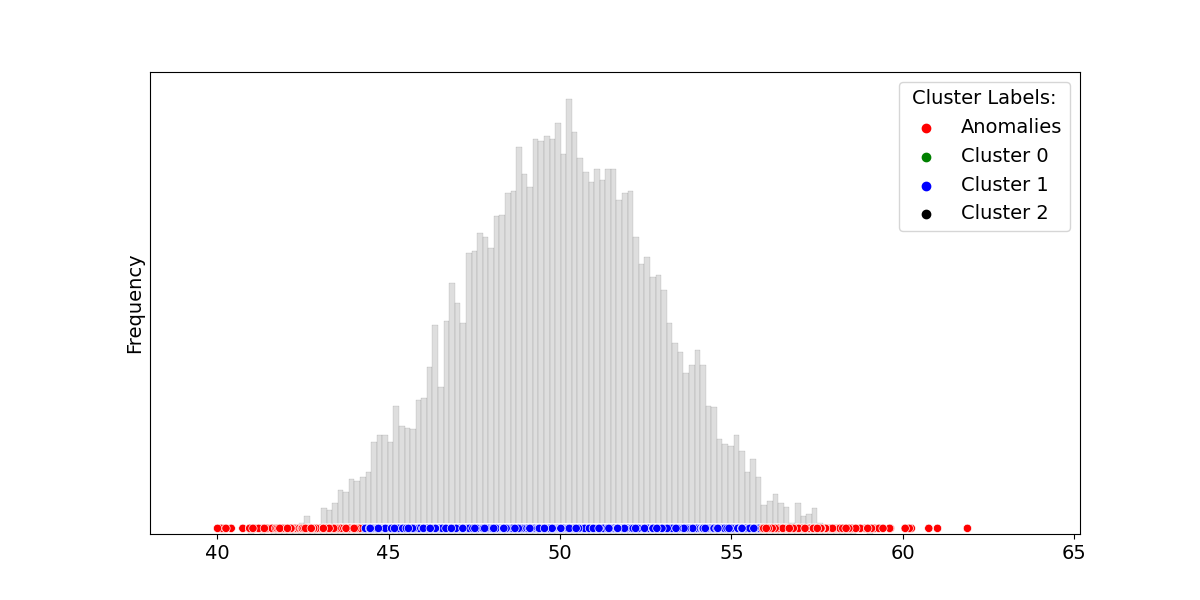}
    \caption{A close up view of cluster 1 from Figure~\ref{fig:1d_nassir_results} showing the identi-
fication of not only anomalies and mislabelled data, but also fringe points
that can be considered rare or unusual in a Gaussian distributed cluster of
points.}
    \label{fig:1d_zoomed}
\end{figure}

The performance of the proposed method is illustrated in Figure~\ref{fig:1d_nassir_results}. The algorithm successfully recovers the three seeded clusters while isolating the anomalous group and individual outliers. A significant advantage is visible in Figure~\ref{fig:1d_zoomed}: the method identifies mislabelled data and fringe points---observations that are rare relative to the cluster mass---as anomalies. This demonstrates a strict adherence to the definition of a cluster as a statistically coherent set rather than a spatial partition.

\begin{figure}
	\includegraphics[width=0.85\linewidth]{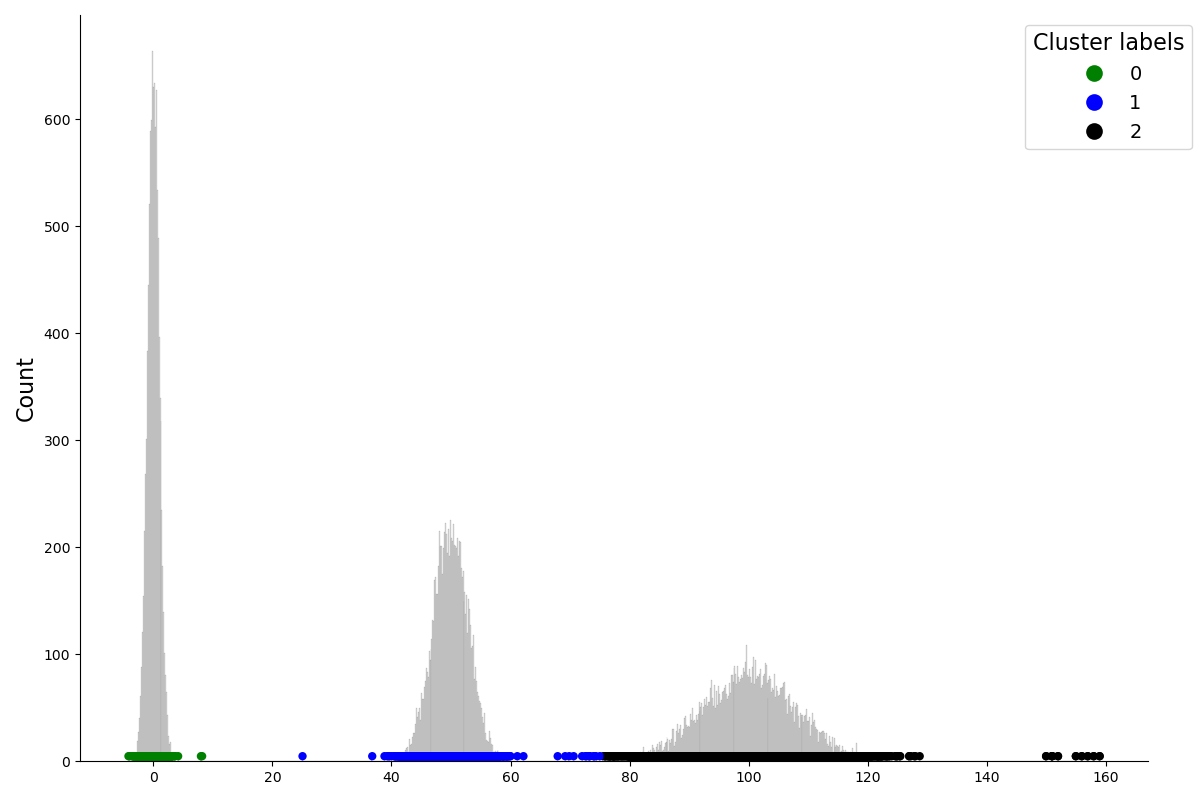}
	\caption{The result of applying $k$-means where all points are put into one of three groups. It hasn't any notion of outliers in the data that do not belong to any cluster.}
	\label{fig:1d_kmeans_results}
\end{figure}

\begin{figure}
	\includegraphics[width=0.85\linewidth]{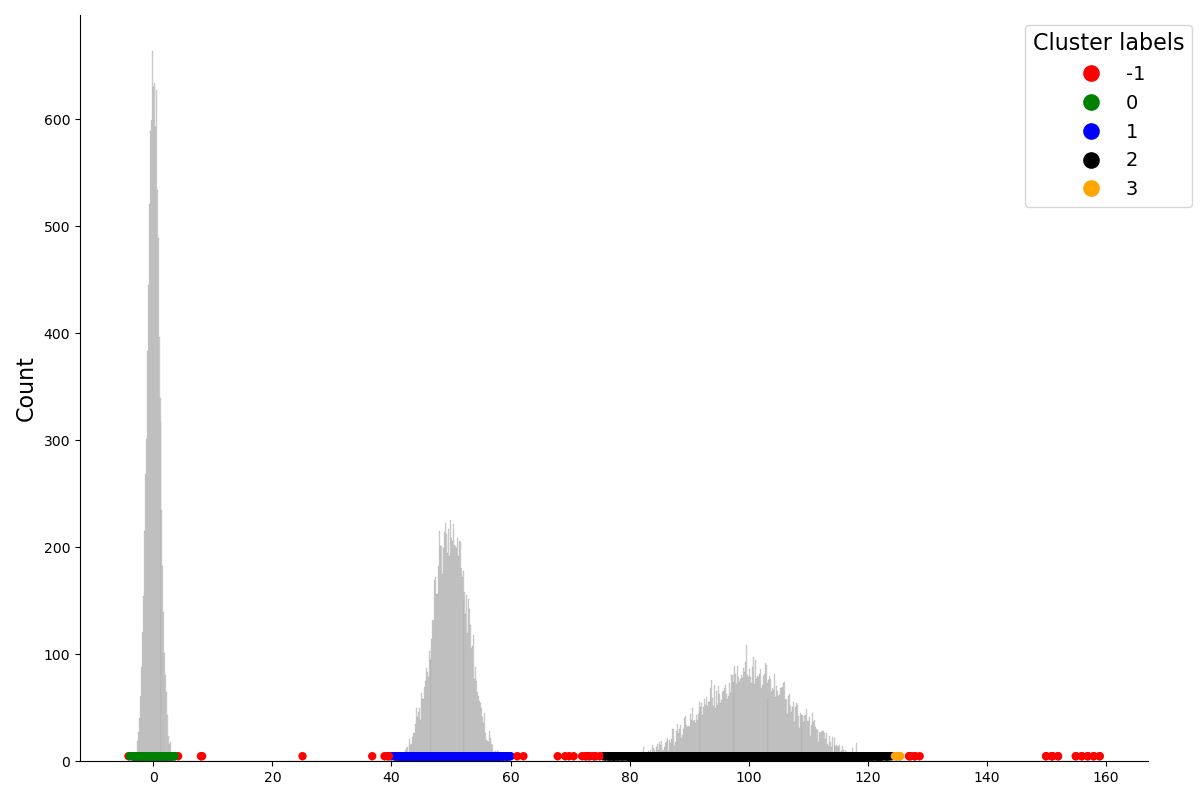}
	\caption{DBSCAN results using its default parameters are shown, where it largely detects the three clusters and also produces anomaly labels for points that were deemed insufficient to be part of a cluster.}
	\label{fig:1d_DBSCAN}
\end{figure}

To provide comparative context, Figure~\ref{fig:1d_kmeans_results} illustrates the results for $k$-means ($k=3$). As a total partitioning algorithm, $k$-means is forced to absorb every outlier and the entire anomalous cluster, which inherently distorts the resulting centroids. While DBSCAN (Figure~\ref{fig:1d_DBSCAN}) performs better by identifying some noise, it fails to provide the nuanced, self-scaling thresholding required to identify fringe points within varying density regions, often resulting in the absorption of Gaussian tails into the cluster core.

\clearpage
\subsection{Two-dimensional Synthetic Data}
\label{section:2d_results}

\begin{figure}[htbp]
    \centering
    \includegraphics[width=0.9\linewidth]{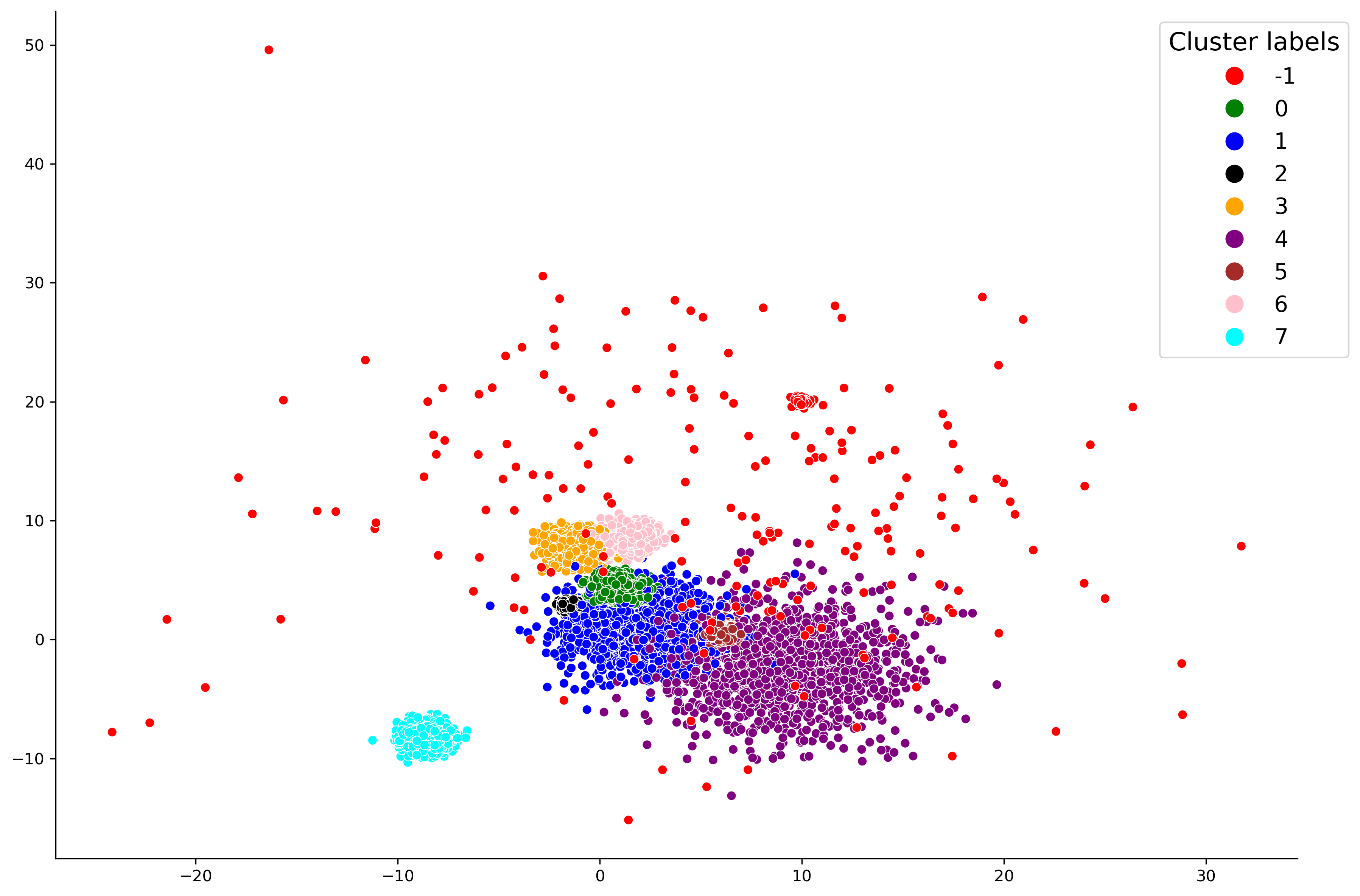}
    \caption{Ground-truth for the \textit{2d\_gauss} dataset featuring eight clusters with heterogeneous scales, alongside isolated noise and a discrete anomalous group.}
    \label{fig:2d_gaussian}
\end{figure}

The next evaluation utilises a two-dimensional dataset, \textit{2d\_gauss}, comprising \num{10300} observations distributed across eight Gaussian clusters with significantly varying scales (standard deviations: $[0.6, 2, 0.2, 0.7, 3, 0.4, 0.6, 0.6]$). As illustrated in Figure~\ref{fig:2d_gaussian}, the set includes scattered isolated anomalies and a concentrated, unseeded group at coordinates $(11, 20)$, representing an emerging cluster for which no prior labels exist.

\begin{figure}[htbp]
    \centering
    \includegraphics[width=0.9\linewidth]{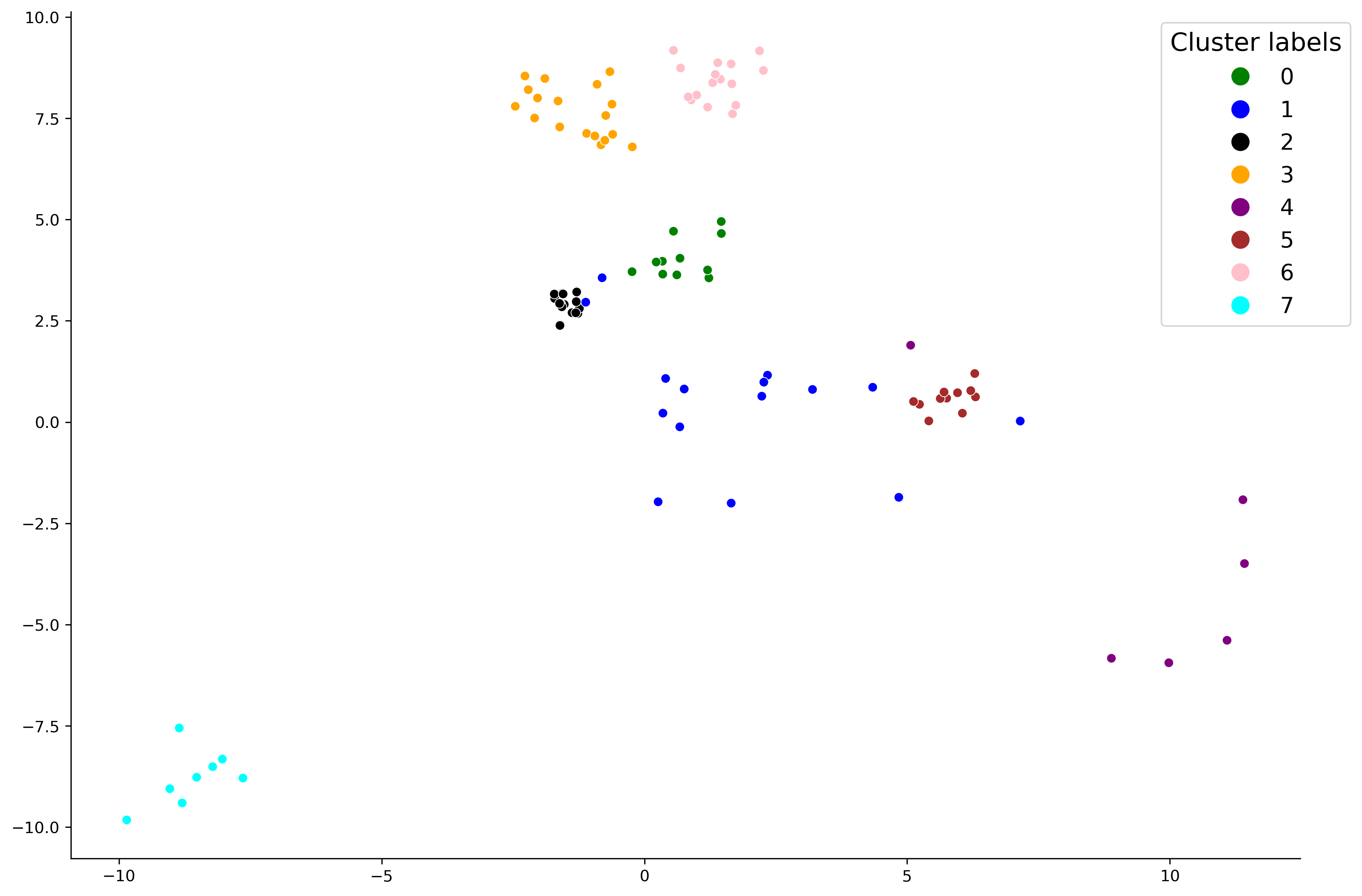}
    \caption{Initial seed distribution (1\% of data) providing the necessary intent to guide the clustering process.}
    \label{fig:2d_gaussian_seeds_only}
\end{figure} 

To guide the expansion, a random sample of approximately 100 points (1\% of total data) was selected as seeds (Figure~\ref{fig:2d_gaussian_seeds_only}). In alignment with our subsequent empirical analysis, $10$--$30$ seeds per cluster are sufficient to anchor the robust marginal medians. Crucially, the iterative nature of the algorithm ensures that the \textit{a-contrario} mechanism remains resilient to noisy supervision; mislabelled or poorly positioned seeds are statistically ejected as anomalies during the refinement phase.

\begin{figure}[htbp]
    \centering
    \includegraphics[width=\linewidth]{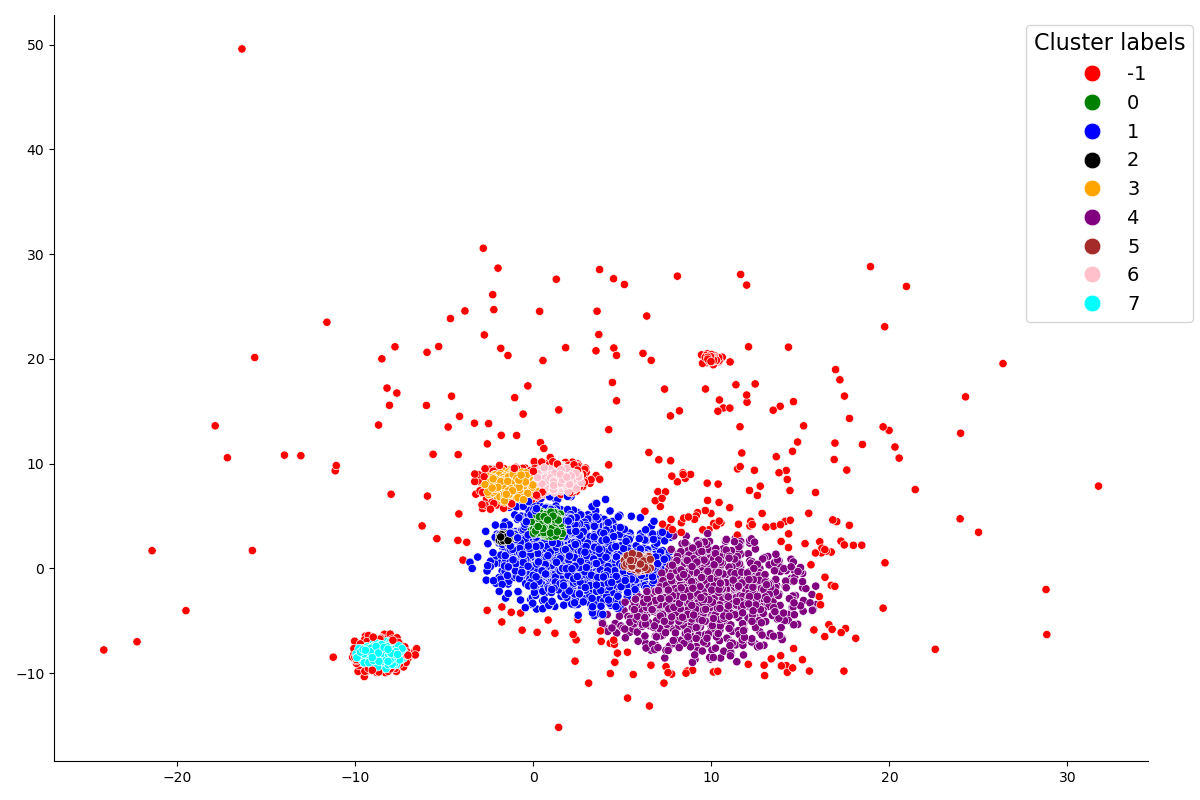}
    \caption{Clustering results of the proposed method. Core structures are recovered, while unseeded groups and isolated noise are correctly identified as anomalies.}
    \label{fig:2d_gaussian_nassir_results}
\end{figure}

\begin{figure}[htbp]
    \centering
    \includegraphics[width=0.9\linewidth]{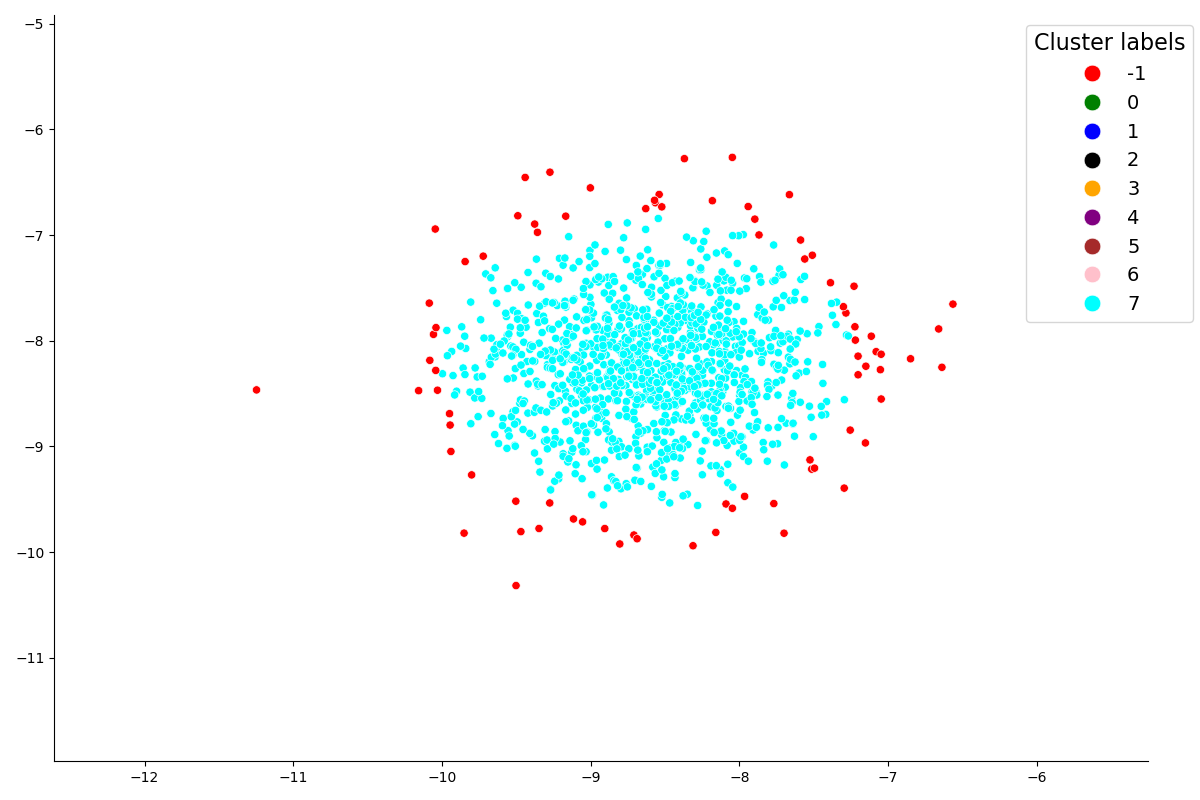}
    \caption{Zoomed view of Cluster 7. The method effectively identifies fringe points as anomalies by applying expectation-based thresholding.}
    \label{fig:2d_gaussian_zoomed}
\end{figure}

The resulting clustering is presented in Figure~\ref{fig:2d_gaussian_nassir_results}. The algorithm successfully delineates all eight clusters, including those that are overlapping or possess disparate densities. Figure~\ref{fig:2d_gaussian_zoomed} highlights a primary advantage of the proposed method: the ability to identify fringe points and outliers relative to the specific local mass of a cluster. Because the expectation $\mathbb{E}$ is calculated relative to the cluster's internal distribution, the threshold naturally adapts to the scale of each group.

\begin{figure}[htbp]
    \centering
    \includegraphics[width=0.9\linewidth]{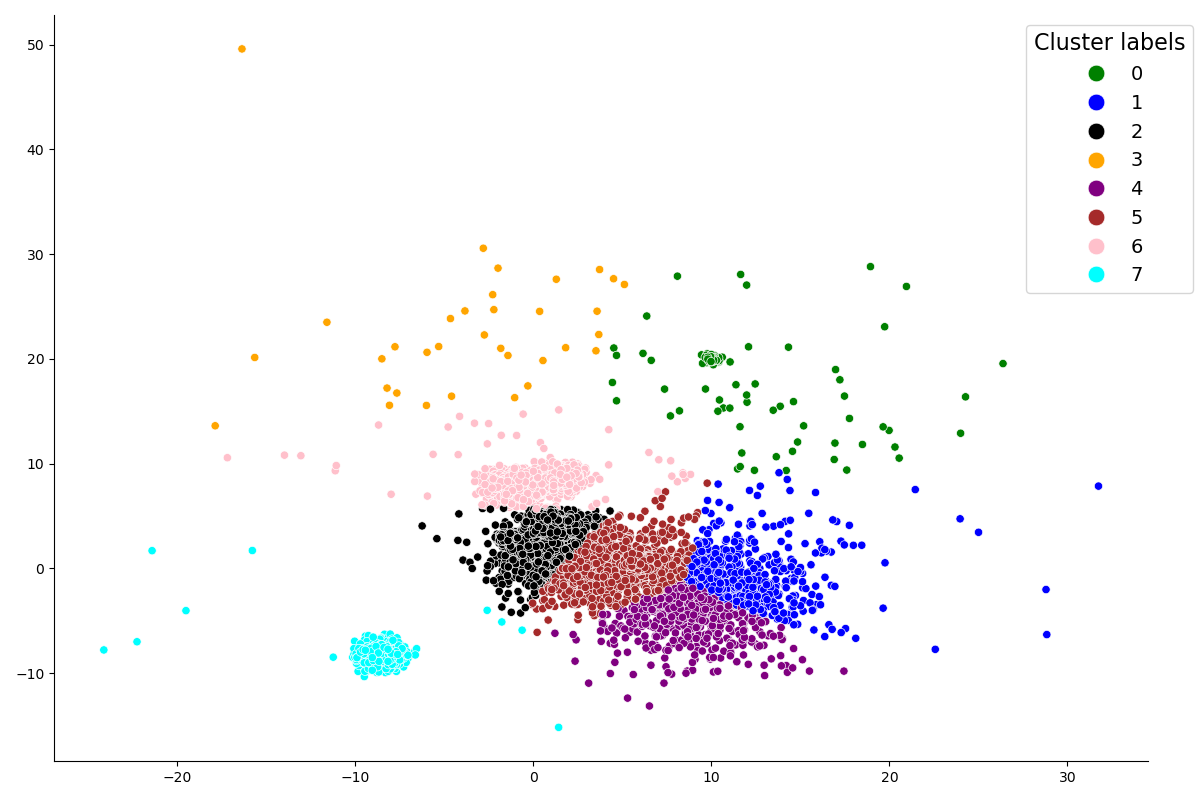}
    \caption{Results for $k$-means. The algorithm fails to accommodate heterogeneous cluster scales and forcibly partitions all anomalies.}
    \label{fig:2d_gaussian_kmeans}
\end{figure}

\begin{figure}[htbp]
    \centering
    \includegraphics[width=0.9\linewidth]{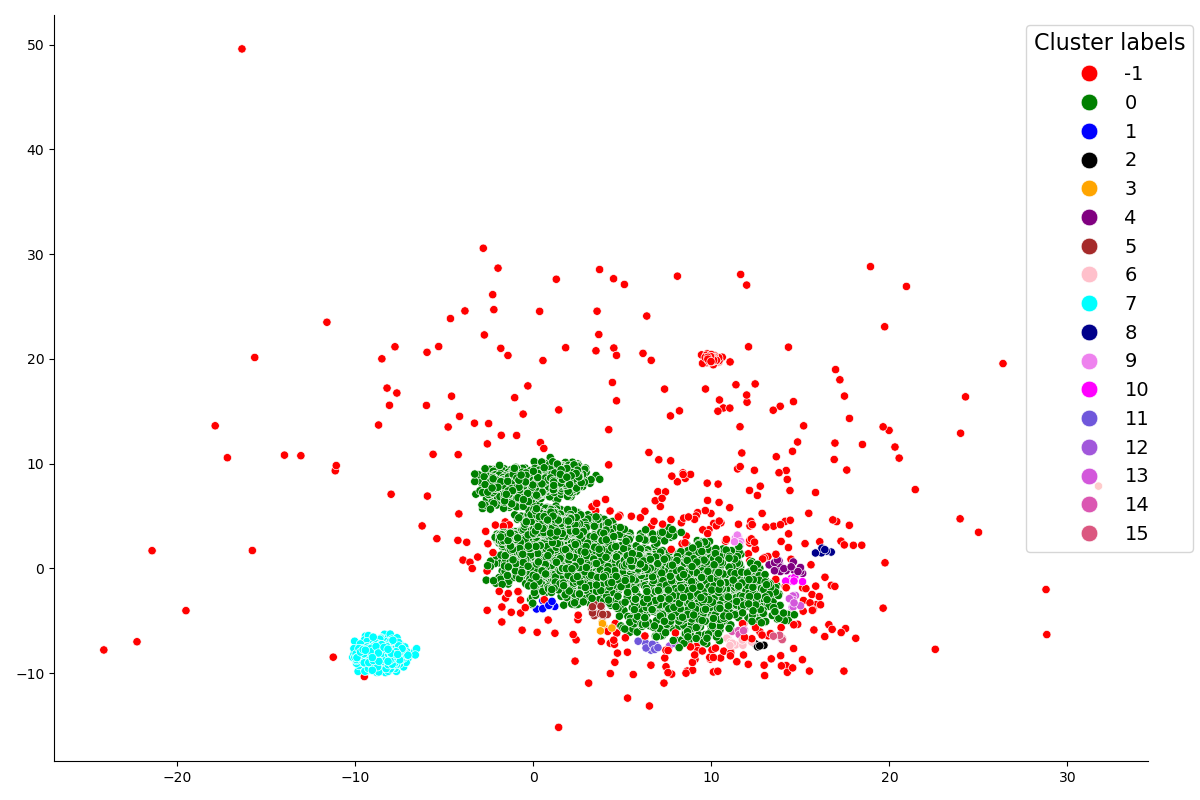}
    \caption{Results for DBSCAN. The lack of adaptive thresholding results in the merging of distinct Gaussian clusters despite identifying some noise.}
    \label{fig:2d_gaussian_dbscan}
\end{figure}

In contrast, $k$-means ($k=8$) produces a mathematically convenient but physically unintuitive partition (Figure~\ref{fig:2d_gaussian_kmeans}). By failing to accommodate varying scales, it merges distinct structures and forcibly assigns anomalies to the nearest centroid. Similarly, DBSCAN (Figure~\ref{fig:2d_gaussian_dbscan}) exhibits the ``chaining effect'', collapsing several distinct clusters into a single mass due to its reliance on a global density parameter. 

\begin{figure}[htbp]
    \centering
    \includegraphics[width=0.9\linewidth]{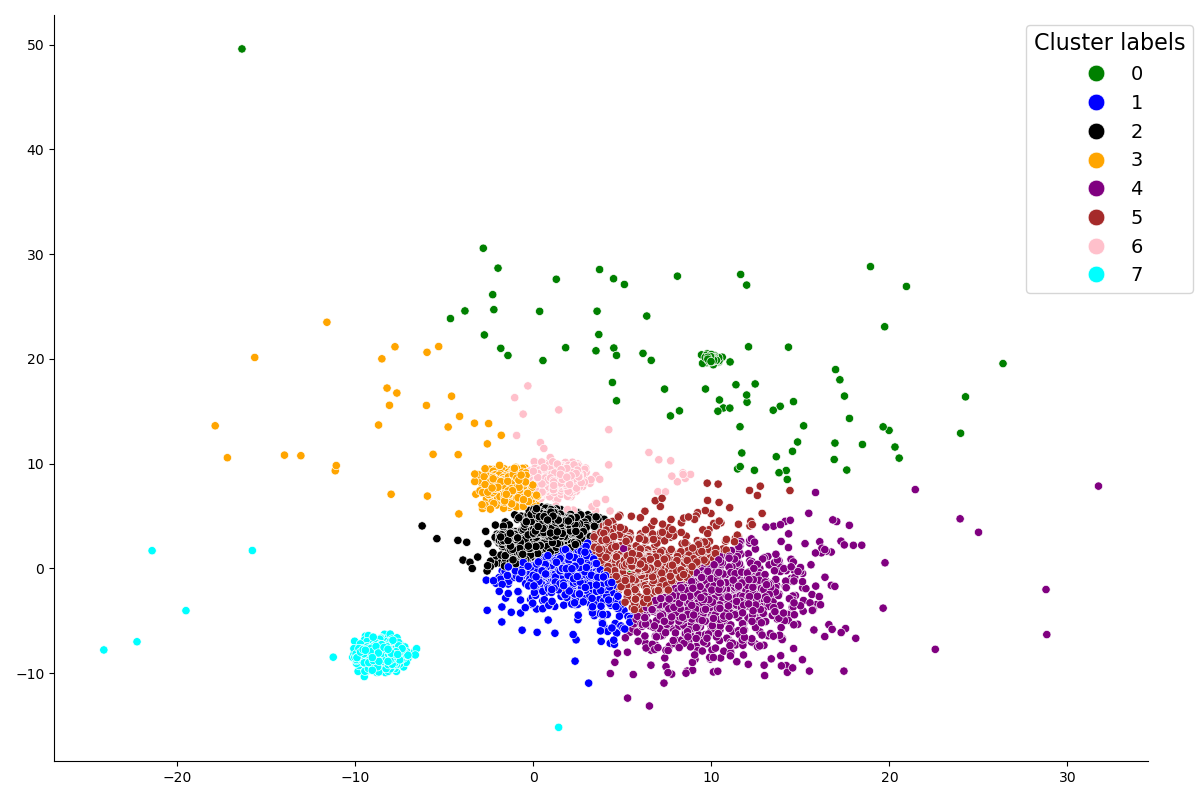}
    \caption{Results for COP-KMeans. Despite seed constraints, the algorithm inherits the partitioning flaws of the $k$-means objective.}
    \label{fig:COPkmeans}
\end{figure}

Perhaps most significantly, semi-supervised baselines like COP-KMeans (Figure~\ref{fig:COPkmeans}) fail to overcome these fundamental geometric limitations. Despite the inclusion of seed constraints, the underlying partitioning objective remains vulnerable to outliers and heterogeneous scales. This indicates that augmenting a distance-to-centroid model with labels is insufficient to achieve the robust boundary definition provided by the proposed clustering-by-exclusion framework, which treats outliers as a structural necessity rather than a secondary nuisance.

\clearpage
\subsection{Multi-dimensional Real-world Benchmarks}
\label{section:realworld_results}

This section evaluates the proposed framework on a diverse suite of real-world benchmarks, ranging from low-dimensional UCI repositories to high-dimensional text and image corpora. These datasets were selected to represent varied structural characteristics, heterogeneous densities, and varying levels of intrinsic complexity \cite{GonzalezAlmagroPPCG25}. Quantitative performance is assessed using five standard external validation metrics: Purity, V-Measure, Normalised Mutual Information (NMI), Adjusted Rand Index (ARI), and the Fowlkes--Mallows Index (FMI). Consistent with our established methodology, the main benchmark tables report
retained-point metrics: external validation scores are computed on observations
with non-anomaly ground-truth labels and non-rejected predicted labels. This
assesses the quality of the admitted cluster assignments without conflating it
with anomaly or rejection behaviour, which is reported separately through
rejection rates and the auxiliary full-dataset evaluation.

Table~\ref{tab:datasets} summarises the dataset characteristics, including the number of observations ($n$), feature dimensionality ($Dim$), and class counts. In the \textit{yeast} and \textit{shuttle} datasets, minority classes were treated as anomalies due to their lack of a coherent statistical structure relative to the dominant groups; this resulted in a reduction from 10 and 7 classes to 4 and 3, respectively. For these datasets this follows common practice in anomaly-detection benchmarking, where the minority or rare classes are treated as outliers; however, in the present work it should be understood as a task-specific relabelling adopted to align the evaluation with the anomaly-aware clustering formulation, rather than as a replacement of the original benchmark annotations. For high-dimensional datasets ($Dim > 30$), such as \textit{ionosphere} and the \textit{MNIST} digits, we applied the Uniform Manifold Approximation and Projection (UMAP) algorithm \cite{mcinnes2018umap} to reduce dimensionality to 10. This reduction was applied uniformly across all baselines to ensure the evaluation isolates clustering logic rather than feature engineering capabilities. Accordingly, the resulting experiments should be interpreted as evaluations in embedding space rather than as a complete characterisation of clustering quality in the original raw feature space. An exception was made for the \textit{cover\_type} dataset, which was retained in its original 54-dimensional form to serve as a rigorous large-scale benchmark for scalability.

For the text benchmark, we use a six-category subset of the 20
Newsgroups corpus; the feature count reflects the dimensionality following
lemmatisation, TF-IDF vectorisation, and subsequent UMAP reduction.
Table~\ref{tab:datasets} also specifies the percentage of labelled data
utilised for the semi-supervised baselines. While smaller datasets required a
slightly higher proportion of seeds to guide the grouping effectively, the
average number of labels per cluster remained consistently low across the
entire suite. All datasets and experimental configurations are accessible in
the accompanying repository; benchmarks were conducted on a MacBook M1 Pro with
16~GB of RAM.

\begin{table}[htbp]
\centering
\caption{Summary of benchmark datasets used in the evaluation.}
\label{tab:datasets}
\resizebox{\linewidth}{!}{%
\begin{tabular}{lrrrrr}
\toprule
\textbf{Dataset} & \textbf{n} & \textbf{Dim} & \textbf{\#Classes} & \textbf{\% Labelled} & \textbf{\#Lbl/Clust.} \\
\midrule
1d\_gauss           & 17,528   & 1   & 3   & 0.2  & 12  \\
2d\_gauss           & 10,300   & 2   & 8   & 1    & 13  \\
6Newsgroups\_UMAP10 & 10,496   & 10  & 6   & 1    & 17  \\
MNIST\_UMAP10       & 1,797    & 10  & 10  & 5    & 9   \\
banknote            & 1,372    & 4   & 2   & 2    & 13  \\
breast\_cancer      & 569      & 30  & 2   & 7    & 20  \\
cover\_type         & 581,012  & 54  & 7   & 0.02 & 15 \\
glass               & 214      & 9   & 6   & 30   & 10  \\
ionosphere\_UMAP10  & 351      & 10  & 2   & 10   & 17  \\
iris                & 150      & 4   & 3   & 20   & 10  \\
land\_mines         & 338      & 3   & 5   & 30   & 20  \\
pendigits           & 10,992   & 16  & 10  & 2.5  & 27  \\
seeds               & 210      & 7   & 3   & 20   & 14  \\
shuttle             & 43,500   & 9   & 3   & 0.2  & 29  \\
wine                & 178      & 13  & 3   & 30   & 17  \\
yeast               & 1,484    & 8   & 4  & 5     & 18   \\
\bottomrule
\end{tabular}
}
\end{table}

\subsection{Performance Analysis and Benchmarking}
\label{subsection:performance_analysis}

Tables~\ref{tab:purity}--\ref{tab:runtime} compare the proposed method with the
benchmark methods under the retained-point evaluation protocol. The method is
among the best-performing approaches across the reported metrics on
\textit{MNIST\_UMAP10}, \textit{6Newsgroups\_UMAP10}, and
\textit{ionosphere\_UMAP10}, and it successfully completes the large-scale
\textit{cover\_type} benchmark where several baselines do not
(Table~\ref{tab:runtime}). Across the full suite, however, the method is not
uniformly dominant: its strongest relative performance occurs on datasets where
the chosen representation yields compact, seed-defined structures, while on
other datasets baselines better matched to the inlier class geometry achieve
higher external scores.

This pattern follows directly from the difference in modelling objective. The
proposed algorithm is not designed to optimise complete-assignment accuracy; it
is designed to recover statistically coherent seeded clusters while leaving
unsupported observations unassigned. LabelSpreading is a particularly strong
competitor when the seed labels propagate cleanly through the similarity graph,
while GMM is strong on low-dimensional datasets such as \textit{iris},
\textit{wine}, and \textit{breast\_cancer}, where the true number of clusters is
supplied and the class structure is well approximated by parametric components.
In such settings, the proposed method should be viewed as a robust
semi-supervised alternative rather than a replacement for methods whose
assumptions closely match the benchmark labels.

The recent baselines add further context. DipNSub, a subspace/projection
method, is highly competitive on the synthetic Gaussian benchmarks and performs
well on some individual real-world datasets, such as \textit{banknote} and in
purity on \textit{MNIST\_UMAP10}. However, its performance varies substantially
across the heterogeneous real-world suite, with weaker ARI/NMI/FMI results on
several tabular datasets and on the embedded MNIST, 6 Newsgroups, and
ionosphere benchmarks compared with the proposed method. DipEnc generally
follows the behaviour of the other deep clustering baselines, giving reasonable
results on several datasets but not matching the strongest retained-point
performance on the embedded image and text benchmarks. This comparison is useful because these methods address representation and
subspace structure, whereas the proposed method adds a statistical
admission/rejection layer on top of the chosen representation.

The empirical runtimes reported in Table~\ref{tab:runtime} are consistent with
the $O(tKND)$ complexity of the proposed framework and its linear dependence on
$N$ and $D$ for fixed $K$ and $t$. On \textit{cover\_type} ($n \approx
581,000$), several baselines---indicated by dashed lines---failed to converge,
encountered memory exhaustion, or exceeded the practical computational budget,
whereas the proposed method completed successfully. On smaller datasets, the
current implementation shows higher relative overhead due to the iterative
admission and ejection phases. This reflects the additional cost of explicit
outlier rejection rather than an inherent limitation of the underlying
complexity; profiling indicates the overhead is primarily attributable to
redundant preprocessing in the current Python implementation, and substantial
speedups are achievable through lower-level optimisation.

Because retained-point evaluation measures the quality of admitted assignments,
it should be read together with coverage. We therefore add two auxiliary
full-dataset views. Table~\ref{tab:full_eval_proposed} reports full-dataset
metrics for the proposed method, where ground-truth outliers and rejected
observations are retained during evaluation as an additional
anomaly/unassigned label, and rejection counts are reported explicitly as
$n/N$. Table~\ref{tab:full_eval_compare} provides a compact full-dataset
comparison across methods using ARI and NMI. For methods without an explicit rejection mechanism, this full-dataset view is
their standard all-observation evaluation. These scores can still differ from
the retained-point tables, because retained-point evaluation excludes
ground-truth anomaly labels and method-specific rejected predictions; methods such as DBSCAN may also differ because they can assign observations to a noise label.

Comparing these auxiliary results with the retained-point scores in
Tables~\ref{tab:purity}--\ref{tab:fmi} makes the coverage--quality trade-off
explicit. When rejection rates are negligible or modest, the retained-point and
full-dataset scores are similar. Larger differences occur on the highest
rejection-rate datasets---\textit{MNIST\_UMAP10},
\textit{6Newsgroups\_UMAP10}, \textit{iris}, and
\textit{ionosphere\_UMAP10}---where rejection ranges from 18\% to 27\%. Three
of these high-rejection cases are also among the strongest retained-point
results, showing that high admitted-cluster quality and a non-negligible
coverage cost can occur together. Conversely, a few datasets, including
\textit{shuttle}, \textit{yeast}, and \textit{banknote}, show modest ARI/NMI
gains under the full-dataset view, indicating that the rejected/unassigned
label can sometimes align with minority or anomalous structure. Overall, the
auxiliary tables clarify when the retained-point conclusions transfer directly
to the full dataset and when they should instead be read as high-quality
cluster-core results obtained at a measurable coverage cost.

Sensitivity to the supervision level, greedy ordering rule, numerical
precision, and representation choice is examined in the following additional
analyses.

\begin{table}[htbp]
  \centering
  \caption{Purity metric results averaged over ten runs.}
  \label{tab:purity}
  \resizebox{\linewidth}{!}{\input{tables/Purity.tex}}
\end{table}

\begin{table}[htbp]
  \centering
  \caption{V-Measure metric results averaged over ten runs.}
  \label{tab:vmeasure}
  \resizebox{\linewidth}{!}{\input{tables/V-Measure.tex}}
\end{table}

\begin{table}[htbp]
  \centering
  \caption{Normalised Mutual Information (NMI) results averaged over ten runs.}
  \label{tab:nmi}
  \resizebox{\linewidth}{!}{\input{tables/NMI.tex}}
\end{table}

\begin{table}[htbp]
  \centering
  \caption{Adjusted Rand Index (ARI) results averaged over ten runs.}
  \label{tab:ari}
  \resizebox{\linewidth}{!}{\input{tables/ARI.tex}}
\end{table}

\begin{table}[htbp]
  \centering
  \caption{Fowlkes--Mallows Index (FMI) results averaged over ten runs.}
  \label{tab:fmi}
  \resizebox{\linewidth}{!}{\input{tables/FMI.tex}}
\end{table}

\begin{table}[htbp]
  \centering
  \caption{Runtime (s) comparison. Dashed lines indicate non-convergence or memory exhaustion.}
  \label{tab:runtime}
  \resizebox{\linewidth}{!}{\input{tables/runtime_s.tex}}
\end{table}

\begin{table}[htbp]
  \centering
  \caption{Full-dataset evaluation of the proposed method. Rejected observations are retained as an additional anomaly/unassigned label during metric computation, and rejection counts are reported explicitly as $n/N$.}
  \label{tab:full_eval_proposed}
  \resizebox{\linewidth}{!}{\input{tables/full_eval_proposed.tex}}
\end{table}

\begin{table}[htbp]
  \centering
  \caption{Compact full-dataset comparison across methods using ARI and NMI.}
  \label{tab:full_eval_compare}
  \resizebox{\linewidth}{!}{\input{tables/full_eval_compare.tex}}
\end{table}

\subsection{Additional Analyses}

\subsubsection{Sensitivity to Supervision}

\begin{figure}[htbp]
    \centering
    \includegraphics[width=0.9\textwidth]{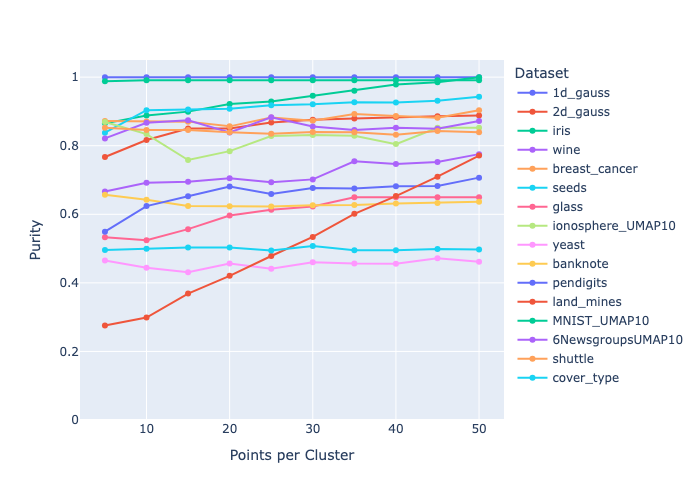}
    \caption{Purity scores of the proposed method as a function of the number of seeds per cluster at initialisation.}
    \label{fig:points_per_cluster_purity}
\end{figure}

\begin{figure}[htbp]
    \centering
    \includegraphics[width=0.9\textwidth]{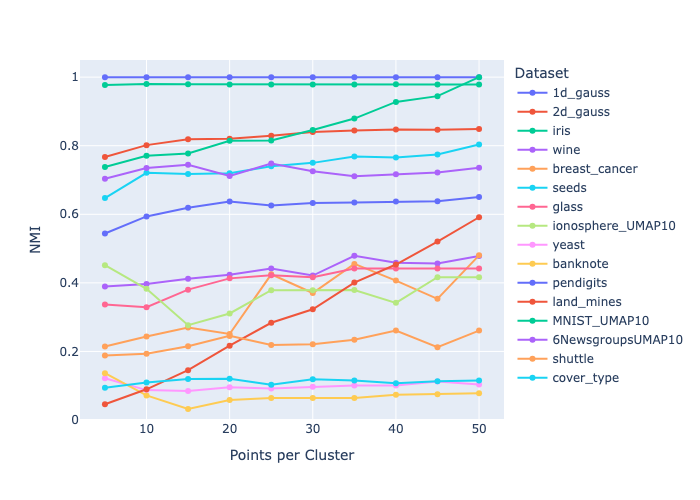}
    \caption{Normalised Mutual Information (NMI) scores as a function of the number of seeds per cluster.}
    \label{fig:nmi_vs_points_per_cluster}
\end{figure}

The relationship between the volume of supervision and clustering quality is illustrated in Figures~\ref{fig:points_per_cluster_purity} and \ref{fig:nmi_vs_points_per_cluster}, which plot Purity and NMI scores against the number of seeds per cluster. These results represent the mean performance over ten independent runs with randomised seed selection. For most datasets, the curves flatten relatively quickly, and typically $10$--$30$ seeds per cluster are sufficient to obtain stable, high-quality groupings, although \textit{land\_mines} continues to benefit more steadily from additional supervision across the tested range. Overall, however, the results still suggest that the method generalises effectively from a small set of representative examples, making it practical for domains where labelling costs are prohibitive.

\subsubsection{Greedy Ordering Ablation}
\label{subsection:overlap_ablation}

The overlap-resolution rule used in Algorithm~\ref{algorithm:Clustering} is a
heuristic ordering rule rather than a quantity derived from the
\textit{a-contrario} model. Because the admission phase is greedy, overlap is
resolved by the order in which clusters claim currently unassigned admissible
points, rather than by an explicit post-hoc tie. If a diffuse or noisy cluster
is processed first, it may claim observations that would also be admissible to
a more compact cluster. The compactness/SSE rule is therefore used to process
tighter clusters first.

To assess this design choice, we compare the current deterministic
compactness/SSE ordering with random greedy priority, keeping all other settings
fixed. For random priority, the cluster processing order is randomly permuted,
and results are reported as mean $\pm$ standard deviation over ten random-order
trials.

\begin{table}[htbp]
\centering
\caption{Ablation of greedy cluster-ordering rules. Compactness/SSE is deterministic; random priority reports mean $\pm$ standard deviation over repeated random cluster-order trials with the same seed set.}
\label{tab:overlap_ablation}
\begin{adjustbox}{width=\linewidth}
\begin{tabular}{llcccc}
\toprule
Dataset & Ordering rule & ARI & NMI & Purity & Rej. (\%) \\
\midrule
\multirow{2}{*}{2d\_gauss} & Compactness/SSE & 0.78 & 0.84 & 0.89 & 6.25 \\
 & Random priority & 0.74 $\pm$ 0.06 & 0.84 $\pm$ 0.02 & 0.84 $\pm$ 0.05 & 5.55 $\pm$ 0.37 \\
\multirow{2}{*}{breast\_cancer} & Compactness/SSE & 0.62 & 0.49 & 0.90 & 6.85 \\
 & Random priority & 0.35 $\pm$ 0.24 & 0.32 $\pm$ 0.14 & 0.89 $\pm$ 0.01 & 23.53 $\pm$ 19.40 \\
\multirow{2}{*}{wine} & Compactness/SSE & 0.36 & 0.41 & 0.70 & 4.49 \\
 & Random priority & 0.20 $\pm$ 0.10 & 0.28 $\pm$ 0.08 & 0.63 $\pm$ 0.04 & 17.19 $\pm$ 10.19 \\
\multirow{2}{*}{MNIST\_UMAP10} & Compactness/SSE & 0.98 & 0.98 & 0.99 & 21.93 \\
 & Random priority & 0.98 $\pm$ 0.00 & 0.98 $\pm$ 0.00 & 0.99 $\pm$ 0.00 & 21.87 $\pm$ 0.00 \\
\bottomrule
\end{tabular}
\end{adjustbox}
\end{table}

The ablation shows that greedy ordering can materially affect the
coverage--quality trade-off. On \textit{2d\_gauss}, compactness/SSE improves
retained-point ARI and purity relative to the mean random-priority result,
while NMI remains comparable. On \textit{breast\_cancer}, random priority is
substantially less stable: the mean ARI and NMI decrease and the rejection
rate has a large standard deviation, consistent with the fact that reversing
the two-cluster processing order can substantially alter which ambiguous
observations are admitted. On \textit{wine}, compactness/SSE improves the
retained-point scores and reduces rejection relative to the mean random
ordering, while \textit{MNIST\_UMAP10} is insensitive to the ordering rule.
These results indicate that compactness/SSE should not be interpreted as an
optimality guarantee; rather, it provides a deterministic ordering that
prioritises compact clusters and avoids stochastic greedy-order effects.

\subsubsection{Sensitivity to Numerical Precision}
\label{subsection:alpha_sensitivity}

To test whether the numerical precision cap $\alpha$ materially affects the
results, we repeated representative experiments across synthetic, image, text,
and tabular benchmarks for $\alpha \in \{2,3,4,5,6\}$. Because $\alpha$ is used
only to integerise continuous distances before applying the fixed
$\mathbb{E}<1$ decision rule, we report the range of each metric across the
sweep rather than treating $\alpha$ as a tuned hyperparameter. Specifically,
each $\Delta$ value is computed as the maximum score over the tested
$\alpha$ values minus the corresponding minimum score.

\begin{table}[htbp]
\centering
\caption{Range of evaluation metrics across $\alpha \in \{2,3,4,5,6\}$, computed as maximum minus minimum over the tested $\alpha$ values. Subscripts $r$ and $f$ denote retained-point and full-dataset evaluation, respectively. Rej. denotes the percentage of observations left unassigned.}
\label{tab:alpha_sensitivity}
\begin{adjustbox}{width=\linewidth}
\begin{tabular}{llrrrrrr}
\toprule
Dataset & Representation & Dim. & $\Delta$ARI$_r$ & $\Delta$NMI$_r$ & $\Delta$ARI$_f$ & $\Delta$NMI$_f$ & $\Delta$Rej. (pp) \\
\midrule
2d Gaussian & Raw features & 2 & 0.0043 & 0.0015 & 0.0036 & 0.0011 & 0.0486 \\
MNIST & UMAP & 10 & 0.0001 & 0.0001 & 0.0212 & 0.0095 & 0.8347 \\
6 Newsgroups & UMAP & 10 & 0.0005 & 0.0021 & 0.0027 & 0.0016 & 0.0850 \\
Wine & Raw features & 13 & 0.0000 & 0.0000 & 0.0000 & 0.0000 & 0.0000 \\
\bottomrule
\end{tabular}
\end{adjustbox}
\end{table}

The results in Table~\ref{tab:alpha_sensitivity} show that retained-point ARI
and NMI are effectively unchanged across the tested precision range. The
largest variation occurs for the full-dataset MNIST scores, where the rejection
rate varies by 0.83 percentage points across the sweep; nevertheless,
retained-point ARI and NMI vary by only 0.0001. This is a useful stress test
for the precision cap because UMAP produces continuous embedding coordinates,
and hence centre-distance values, with multiple decimal places. The limited
variation is consistent with the role of $\alpha$ as a discretisation
precision: coarser settings can merge distance values that differ only beyond
the retained decimal places. These results support treating $\alpha$ as a
fixed numerical precision cap rather than as a clustering decision threshold.
All main experiments therefore use the global default $\alpha=4$, which lies
within the stable range observed in the sensitivity sweep.

\subsubsection{Representation Controls on High-Dimensional Data}
\label{subsection:representation_controls}

To assess whether the proposed method depends specifically on nonlinear UMAP preprocessing, we conducted additional representation-control experiments on the two highest-dimensional benchmarks considered in this study: MNIST and 6 Newsgroups. For MNIST, we compare the original pixel representation, a linear PCA embedding, and the UMAP embedding used in the main benchmark. For 6 Newsgroups, we compare the original TF-IDF representation, a linear TruncatedSVD embedding, and the UMAP embedding used in the main benchmark. TruncatedSVD is used as the sparse-text analogue of PCA, since the TF-IDF matrix is high-dimensional and sparse.

The purpose of this experiment is not to treat dimensionality reduction as an incidental preprocessing detail, but to make explicit how strongly the proposed centre-distance representation depends on the feature space in which distances are computed. The same seed protocol, clustering procedure, and evaluation metrics are used across all representations.

\begin{table}[htbp]
\centering
\caption{Effect of representation on the proposed method for high-dimensional benchmark datasets. Raw high-dimensional features are compared with linear dimensionality reduction and UMAP embeddings. Subscripts $r$ and $f$ denote retained-point and full-dataset evaluation, respectively. Rej. denotes the percentage of observations left unassigned by the proposed method.}
\label{tab:representation_controls}
\begin{adjustbox}{width=\linewidth}
\begin{tabular}{llcccccc}
\toprule
Dataset & Representation & Dim. & ARI$_r$ & NMI$_r$ & ARI$_f$ & NMI$_f$ & Rej. (\%) \\
\midrule
MNIST & Raw pixels & 64 & 0.48 & 0.62 & 0.48 & 0.62 & 0.00 \\
MNIST & PCA & 10 & 0.43 & 0.61 & 0.43 & 0.60 & 0.28 \\
MNIST & UMAP & 10 & 0.98 & 0.98 & 0.64 & 0.80 & 22 \\
\midrule
6 Newsgroups & Raw TF-IDF & 2000 & 0.00 & 0.04 & 0.00 & 0.04 & 2.79 \\
6 Newsgroups & TruncatedSVD & 10 & 0.44 & 0.49 & 0.43 & 0.48 & 1.41 \\
6 Newsgroups & UMAP & 10 & 0.77 & 0.76 & 0.54 & 0.60 & 18 \\
\bottomrule
\end{tabular}
\end{adjustbox}
\end{table}

The results in Table~\ref{tab:representation_controls} quantify the sensitivity of the proposed method to the representation used for high-dimensional data. This sensitivity is substantial. On the digit images, the raw 64-dimensional pixel representation gives moderate retained-point performance (ARI$_r=0.48$, NMI$_r=0.62$), while PCA to 10 dimensions does not improve the result. By contrast, the UMAP representation yields a much stronger retained-point score (ARI$_r=0.98$, NMI$_r=0.98$), albeit with a larger rejection rate and a corresponding full-dataset penalty. The pattern is even clearer for the 6 Newsgroups corpus: raw TF-IDF is ineffective for the present centre-distance formulation (ARI$_r=0.00$, NMI$_r=0.04$), TruncatedSVD provides a partial improvement, and UMAP gives the strongest retained-point result.

These results show that the current implementation should be understood as an
anomaly-aware admission/rejection mechanism applied after a representation has
been chosen, rather than as an image-specific or text-specific clustering
pipeline that is independent of representation quality. This behaviour is
expected, since the algorithm evaluates statistical coherence through distances
to seed-defined cluster centres; in raw pixel or sparse TF-IDF spaces,
Euclidean or cosine distances can be degraded by sparsity, irrelevant
variation, and distance concentration. Linear dimensionality reduction provides
a useful control derived without nonlinear manifold learning, while UMAP tests
the method in a neighbourhood-preserving embedding where local proximity is more
meaningful. Accordingly, the high-dimensional experiments should be interpreted
as evidence for the proposed mechanism in suitable representations, not as
proof that the current centre-based implementation is invariant to arbitrary
raw high-dimensional feature spaces. UMAP is not part of the statistical
thresholding rule itself; it is a representation step preceding the fixed
\textit{a-contrario} decision criterion.

\subsection{Qualitative Analysis of High-Dimensional Corpora}
\label{section:qualitative_analysis}

To further evaluate the practical utility of the proposed framework in complex domains, we examine two high-dimensional benchmarks: the MNIST handwritten digits dataset and a six-category subset of the 20 Newsgroups corpus. These datasets represent distinct challenges in computer vision and natural language processing, characterised by non-linear manifolds and inherently subjective ground-truth labels.

\subsubsection{MNIST Handwritten Digits}
The MNIST dataset presents a significant challenge due to the high dimensionality of raw pixel data and the significant variance in individual handwriting styles. To enhance clustering stability and address the sparsity of the raw feature space, we utilised UMAP \cite{mcinnes2018umap} with a cosine metric to project the data into a 10-dimensional latent space. We note that this preprocessing step may discard information; the results reported here therefore demonstrate performance in a reduced representation where local proximity is more meaningful for distance-based grouping.

\begin{figure}[htbp]
    \centering
    \includegraphics[width=\linewidth]{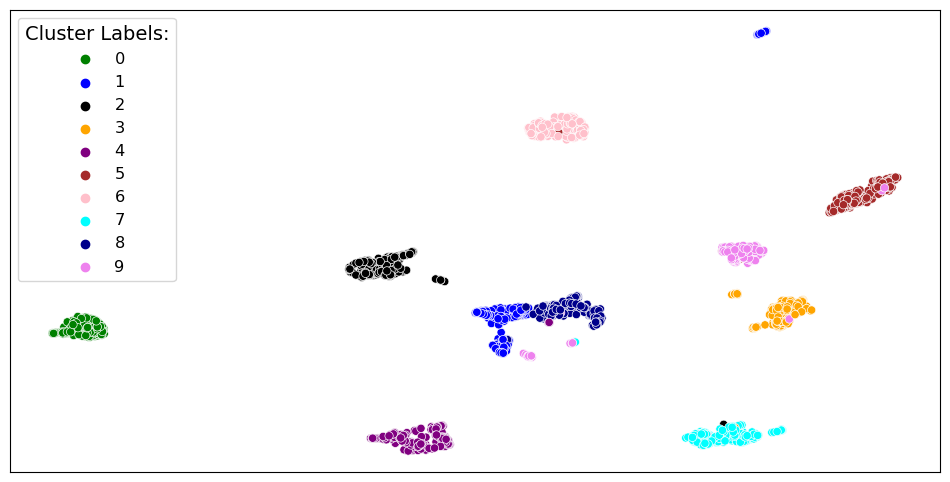}
    \caption{2-dimensional UMAP projection of the MNIST dataset. Points are colour-coded by ground-truth labels, revealing latent ambiguity between digit classes.}
    \label{fig:digits_umap_labelled}
\end{figure}

Figure~\ref{fig:digits_umap_labelled} illustrates the ground-truth distribution. While ten prominent clusters are visible, the presence of smaller, disparate groupings suggests that the benchmark labels may not fully capture all anomaly-relevant structure; digit interpretation is often ambiguous even to human observers, and anomaly definitions can depend on the modelling perspective. For the clustering process, we utilised a 5\% random sample (90 examples) as seeds to initialise the process.

\begin{figure}[htbp]
    \centering
    \includegraphics[width=\linewidth]{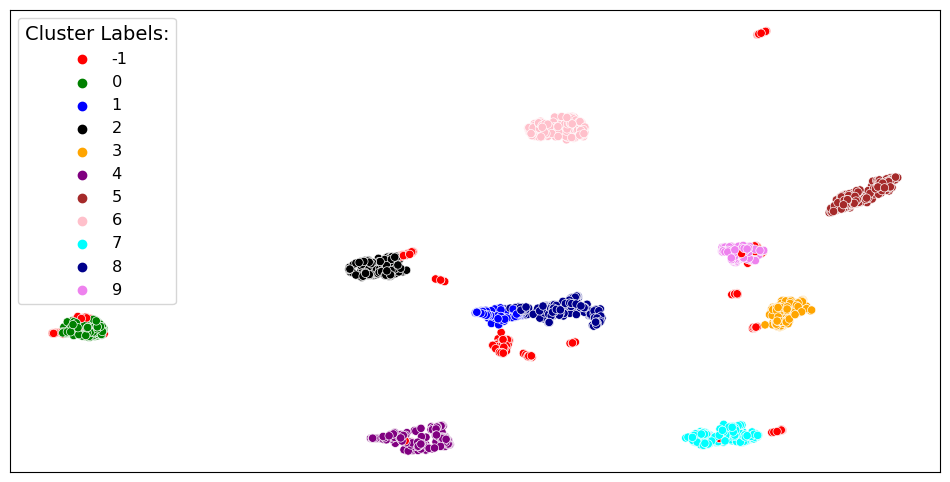}
    \caption{Clustering results on MNIST ($Dim=10$). The algorithm successfully delineates digit classes while relegating ambiguous or rare variations to the anomaly set.}
    \label{fig:digits_umap_nassir}
\end{figure}

The qualitative results, shown in Figure~\ref{fig:digits_umap_nassir}, support the top-tier performance obtained according to the external validation metrics relative to established baselines. Furthermore, a closer examination of the visual results demonstrates that the \textit{clustering-by-exclusion} approach identifies stylistic variations---such as the digit `1' written with a horizontal baseline---as anomalies. By refusing to force these stylistic outliers into the main centroids, the method preserves the statistical integrity of the core clusters and prevents the ``smearing'' of cluster boundaries often observed in partitioning methods.

\subsubsection{20 Newsgroups Corpus}
\begin{figure}[htbp]
    \centering
    \includegraphics[width=\linewidth]{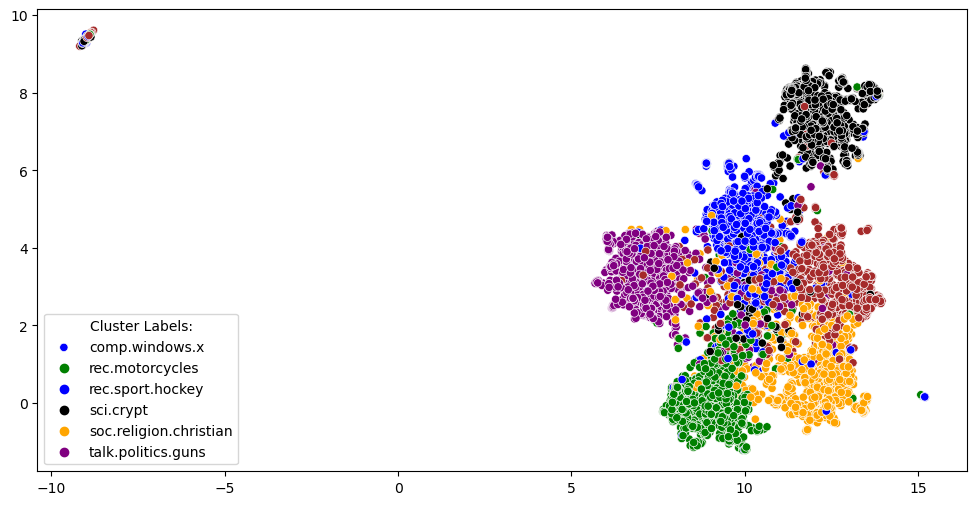}
    \caption{2-dimensional UMAP projection of the 6-category Newsgroup subset coloured by ground-truth labels.}
    \label{fig:20newsgrp_umap_labelled}
\end{figure}

The 20 Newsgroups corpus is a standard benchmark for semantic document clustering. We selected a subset of six distinct categories to assess the algorithm's performance in a high-dimensional TF-IDF space reduced to 10 dimensions via UMAP. Figure~\ref{fig:20newsgrp_umap_labelled} reveals significant overlap between semantic categories and a disparate cluster (top-left) containing examples from multiple ground-truth classes.

\begin{figure}[htbp]
    \centering
    \includegraphics[width=\linewidth]{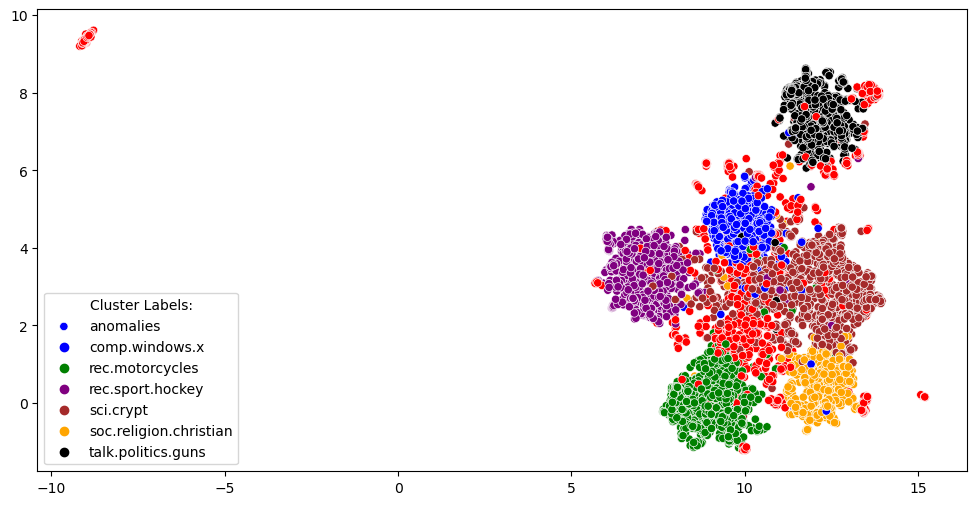}
    \caption{Clustering results on the Newsgroup dataset. The algorithm effectively delineates semantic boundaries, isolating transitional documents as anomalies.}
    \label{fig:20newsgrp_umap_nassir}
\end{figure}

By leveraging a minimal 1\% seed sample, the algorithm successfully recovered the six primary categories (Figure~\ref{fig:20newsgrp_umap_nassir}). A key finding is the algorithm's treatment of transitional documents---those located at the semantic boundaries between topics---which are flagged as anomalies. This behaviour supports that the \textit{a-contrario} mechanism acts as a natural ``safety valve,'' preventing the corruption of cluster centroids by documents with ambiguous or multi-topic signatures.

Table~\ref{table:cluster_keywords} presents the top 10 keywords for each recovered cluster. While the keywords for the six primary clusters align with their respective category labels (e.g., \textit{encryption}, \textit{hockey}, \textit{bible}), the keywords for the anomalous group represent a generic, incoherent mix (e.g., \textit{thanks}, \textit{like}, \textit{com}). This suggests that the rejected points do not collectively behave as a single coherent seventh semantic category. Instead, they appear to form a heterogeneous residual set, potentially comprising transitional documents, unstructured noise, and smaller unseeded subgroups, consistent with their exclusion from the statistically coherent primary clusters.

\begin{table}[htbp]
    \centering
    \caption{Top 10 Keywords for Recovered Clusters and the Anomaly Group.}
    \label{table:cluster_keywords}
    \small 
    \begin{tabular}{|c|p{0.7\textwidth}|} 
    \hline
    \textbf{Category} & \textbf{Keywords} \\
    \hline
    \textbf{Anomalies} & helmet, thanks, edu, just, like, list, mail, com, dog, wa \\
    \hline
    \textbf{comp.windows.x} & use, display, application, program, motif, widget, thanks, file, server, window \\
    \hline
    \textbf{rec.motorcycles} & rider, road, just, like, riding, dod, motorcycle, ride, wa, bike \\
    \hline
    \textbf{rec.sport.hockey} & espn, season, playoff, play, year, player, hockey, wa, team, game \\
    \hline
    \textbf{sci.crypt} & escrow, use, nsa, algorithm, phone, government, encryption, clipper, chip, key \\
    \hline
    \textbf{soc.religion.christian} & christians, bible, people, sin, christ, christian, church, jesus, wa, god \\
    \hline
    \textbf{talk.politics.guns} & just, weapon, government, law, did, right, fbi, people, wa, gun \\
    \hline
    \end{tabular}
\end{table}

\section{Conclusion}
\label{section:conclusion}

This work has presented a formalisation of clustering as the statistical dual of anomaly detection, structured according to Marr's tri-level hypothesis \cite{Mar82}. By linking Gestalt grouping principles with the \textit{a-contrario} framework, we defined a cluster as a maximal subset of observations devoid of internal anomalies relative to a null hypothesis of uniform randomness. This provides a principled clustering criterion in which meaningful structure is identified through statistical unexpectedness rather than through manually tuned decision thresholds alone. On this basis, we developed a semi-supervised \textit{clustering-by-exclusion} algorithm built on the Perception anomaly detection kernel \cite{nassir2021anomaly}. Starting from a small number of seeds, the method iteratively admits statistically coherent points while ejecting inconsistent or fringe observations. In this way, it preserves compact cluster cores without forcing every observation into a partition, and it replaces global hyperparameter search with a small amount of interpretable supervision that expresses the intended clustering resolution. 

The empirical results show that this approach is practically effective across synthetic and real-world benchmarks. Under the low-tuning protocol adopted here, the method recovers strong retained-point cluster quality while maintaining $O(tKND)$ worst-case time complexity, which is linear in the number of observations and dimensions for fixed $K$ and $t$, and scaling to large datasets such as \textit{cover\_type}. The added full-dataset evaluation further clarifies the expected trade-off between coverage and quality: performance changes little when rejection rates are low, but decreases more noticeably when a substantial fraction of observations is explicitly left unassigned. At the same time, the qualitative analyses of \textit{MNIST\_UMAP10} and \textit{6Newsgroups\_UMAP10} indicate that some rejected observations correspond to visually or semantically distinct fringe structure that is not fully represented by the benchmark ground truth, so standard external metrics do not always capture the full anomaly-aware behaviour of the method.

These results should also be interpreted in light of the method's current scope. The present formulation is centre-based and is therefore best suited to compact, approximately spherical groups; it is expected to be less effective when classes are strongly overlapping, genuinely multi-modal, or embedded in spaces where the chosen distance representation does not preserve meaningful local structure. The experiments on UMAP embeddings therefore validate the method in a reduced representation where local proximity is more meaningful, rather than providing a complete characterisation in the original very high-dimensional raw feature spaces.

More broadly, the framework supports an interactive view of clustering in which
seeds express user intent directly, and rejected observations provide a natural
focus for subsequent inspection. This is particularly relevant for large-scale
industrial settings, where exhaustive parameter search and trial-and-error
validation can be costly: expert input is concentrated into representative
seeds and, when needed, the review of rejected observations for possible new
target groups. The current implementation is designed for static batch
datasets; applying the method in streaming or dynamic environments would
require incremental updates of scaling statistics, cluster medians, distance
counts, and admission/rejection decisions. Future work will therefore explore
alternative grouping laws and representations within the same
\textit{a-contrario} framework, null models that remain stable directly in very
high-dimensional sparse spaces, automated procedures for suggesting new seeds
from the rejected set, online or streaming variants of the admission/ejection
mechanism, and further optimisation of the iterative phases.


\bibliography{Bibliography.bib} 
\bibliographystyle{plainnat}

\end{document}

%% file: tables/Purity.tex
\begin{tabular}{lrrrrrrrrrrrr}
\toprule
 & KM & GMM & S-KM & C-KM & COP-KM & Agg & DBSCAN & DEC & DipNSub & DipEnc & LS & Ours \\
dataset &  &  &  &  &  &  &  &  &  &  &  &  \\
\midrule
1d\_gauss & \textbf{1.00} & \textbf{1.00} & \textbf{1.00} & 0.93 & \textbf{1.00} & \textbf{1.00} & \textbf{1.00} & \textbf{1.00} & \textbf{1.00} & \textbf{1.00} & 0.63 & \textbf{1.00} \\
2d\_gauss & 0.81 & 0.88 & 0.84 & 0.81 & 0.77 & 0.78 & 0.52 & 0.84 & 0.86 & 0.67 & 0.77 & \textbf{0.89} \\
6NewsgroupsUMAP10 & 0.74 & 0.74 & 0.78 & 0.74 & 0.74 & 0.72 & 0.34 & 0.74 & 0.82 & 0.74 & 0.76 & \textbf{0.89} \\
MNIST\_UMAP10 & 0.88 & 0.88 & 0.88 & 0.87 & 0.87 & 0.88 & 0.97 & 0.88 & 0.96 & 0.88 & 0.98 & \textbf{0.99} \\
banknote & 0.61 & 0.56 & 0.61 & 0.61 & 0.62 & 0.56 & 0.70 & 0.56 & 0.80 & 0.56 & \textbf{0.89} & 0.62 \\
breast\_cancer & 0.85 & \textbf{0.95} & 0.85 & 0.87 & 0.86 & 0.78 & 0.63 & 0.85 & 0.63 & 0.82 & 0.90 & 0.90 \\
cover\_type & 0.49 & \textbf{0.57} & 0.49 & -- & -- & -- & -- & 0.49 & -- & -- & -- & 0.50 \\
glass & 0.59 & 0.56 & 0.57 & 0.62 & 0.62 & 0.54 & 0.49 & 0.43 & 0.43 & 0.36 & \textbf{0.70} & 0.56 \\
ionosphere\_UMAP10 & 0.70 & 0.69 & 0.70 & 0.70 & 0.72 & 0.70 & 0.64 & 0.68 & \textbf{0.85} & 0.70 & 0.67 & \textbf{0.85} \\
iris & 0.89 & \textbf{0.97} & 0.89 & 0.89 & 0.79 & 0.89 & 0.69 & 0.93 & 0.77 & 0.91 & \textbf{0.97} & 0.89 \\
land\_mines & 0.23 & 0.38 & 0.30 & 0.30 & 0.45 & 0.28 & 0.21 & 0.26 & 0.21 & 0.25 & \textbf{0.52} & 0.41 \\
pendigits & 0.75 & 0.71 & 0.77 & 0.77 & 0.71 & 0.72 & 0.10 & 0.75 & 0.60 & 0.72 & \textbf{0.96} & 0.69 \\
seeds & 0.90 & \textbf{0.93} & 0.90 & 0.90 & 0.91 & 0.89 & 0.53 & 0.86 & 0.33 & 0.67 & 0.91 & 0.90 \\
shuttle & 0.79 & 0.79 & 0.83 & 0.79 & -- & -- & 0.79 & 0.79 & \textbf{1.00} & 0.79 & 0.93 & 0.84 \\
wine & 0.70 & \textbf{0.85} & 0.70 & 0.70 & 0.78 & 0.70 & 0.40 & 0.70 & 0.40 & 0.67 & 0.77 & 0.70 \\
yeast & 0.49 & 0.44 & 0.50 & 0.49 & 0.49 & 0.47 & 0.36 & 0.47 & 0.43 & 0.51 & \textbf{0.52} & 0.44 \\
\bottomrule
\end{tabular}

%% file: tables/V-Measure.tex
\begin{tabular}{lrrrrrrrrrrrr}
\toprule
 & KM & GMM & S-KM & C-KM & COP-KM & Agg & DBSCAN & DEC & DipNSub & DipEnc & LS & Ours \\
dataset &  &  &  &  &  &  &  &  &  &  &  &  \\
\midrule
1d\_gauss & \textbf{1.00} & \textbf{1.00} & \textbf{1.00} & 0.80 & \textbf{1.00} & \textbf{1.00} & \textbf{1.00} & \textbf{1.00} & \textbf{1.00} & \textbf{1.00} & 0.16 & \textbf{1.00} \\
2d\_gauss & 0.87 & \textbf{0.92} & 0.83 & 0.83 & 0.82 & 0.85 & 0.72 & 0.85 & 0.85 & 0.80 & 0.71 & 0.84 \\
6NewsgroupsUMAP10 & 0.61 & 0.61 & 0.62 & 0.60 & 0.60 & 0.61 & 0.29 & 0.62 & 0.57 & 0.61 & 0.51 & \textbf{0.76} \\
MNIST\_UMAP10 & 0.90 & 0.90 & 0.90 & 0.87 & 0.89 & 0.90 & 0.93 & 0.90 & 0.87 & 0.90 & 0.95 & \textbf{0.98} \\
banknote & 0.03 & 0.01 & 0.03 & 0.03 & 0.03 & 0.00 & 0.17 & 0.01 & 0.33 & 0.01 & \textbf{0.57} & 0.01 \\
breast\_cancer & 0.46 & \textbf{0.71} & 0.46 & 0.50 & 0.48 & 0.32 & 0.00 & 0.46 & 0.00 & 0.41 & 0.52 & 0.49 \\
cover\_type & 0.07 & \textbf{0.17} & 0.07 & -- & -- & -- & -- & 0.04 & -- & -- & -- & 0.11 \\
glass & 0.43 & 0.40 & 0.40 & 0.36 & 0.45 & 0.39 & 0.31 & 0.14 & 0.24 & 0.06 & \textbf{0.51} & 0.36 \\
ionosphere\_UMAP10 & 0.12 & 0.11 & 0.12 & 0.12 & 0.16 & 0.12 & 0.10 & 0.11 & 0.23 & 0.12 & 0.11 & \textbf{0.42} \\
iris & 0.74 & \textbf{0.90} & 0.74 & 0.76 & 0.63 & 0.77 & 0.60 & 0.81 & 0.39 & 0.81 & 0.88 & 0.77 \\
land\_mines & 0.01 & 0.20 & 0.10 & 0.12 & 0.18 & 0.08 & 0.00 & 0.01 & 0.00 & 0.01 & \textbf{0.24} & 0.20 \\
pendigits & 0.69 & 0.68 & 0.69 & 0.69 & 0.68 & 0.73 & 0.00 & 0.74 & 0.38 & 0.71 & \textbf{0.91} & 0.64 \\
seeds & 0.69 & \textbf{0.76} & 0.69 & 0.69 & 0.73 & 0.73 & 0.29 & 0.65 & 0.00 & 0.55 & 0.71 & 0.69 \\
shuttle & 0.00 & 0.00 & 0.26 & 0.16 & -- & -- & 0.00 & 0.08 & 0.16 & 0.12 & \textbf{0.64} & 0.21 \\
wine & 0.43 & \textbf{0.58} & 0.43 & 0.43 & 0.50 & 0.42 & 0.00 & 0.43 & 0.00 & 0.38 & 0.45 & 0.41 \\
yeast & 0.12 & 0.11 & 0.16 & 0.12 & 0.13 & 0.13 & 0.00 & 0.09 & 0.03 & 0.16 & \textbf{0.19} & 0.08 \\
\bottomrule
\end{tabular}

%% file: tables/NMI.tex
\begin{tabular}{lrrrrrrrrrrrr}
\toprule
 & KM & GMM & S-KM & C-KM & COP-KM & Agg & DBSCAN & DEC & DipNSub & DipEnc & LS & Ours \\
dataset &  &  &  &  &  &  &  &  &  &  &  &  \\
\midrule
1d\_gauss & \textbf{1.00} & \textbf{1.00} & \textbf{1.00} & 0.80 & \textbf{1.00} & \textbf{1.00} & \textbf{1.00} & \textbf{1.00} & \textbf{1.00} & \textbf{1.00} & 0.16 & \textbf{1.00} \\
2d\_gauss & 0.87 & \textbf{0.92} & 0.83 & 0.83 & 0.82 & 0.85 & 0.72 & 0.85 & 0.85 & 0.80 & 0.71 & 0.84 \\
6NewsgroupsUMAP10 & 0.61 & 0.61 & 0.62 & 0.60 & 0.60 & 0.61 & 0.29 & 0.62 & 0.57 & 0.61 & 0.51 & \textbf{0.76} \\
MNIST\_UMAP10 & 0.90 & 0.90 & 0.90 & 0.87 & 0.89 & 0.90 & 0.93 & 0.90 & 0.87 & 0.90 & 0.95 & \textbf{0.98} \\
banknote & 0.03 & 0.01 & 0.03 & 0.03 & 0.03 & 0.00 & 0.17 & 0.01 & 0.33 & 0.01 & \textbf{0.57} & 0.01 \\
breast\_cancer & 0.46 & \textbf{0.71} & 0.46 & 0.50 & 0.48 & 0.32 & 0.00 & 0.46 & 0.00 & 0.41 & 0.52 & 0.49 \\
cover\_type & 0.07 & \textbf{0.17} & 0.07 & -- & -- & -- & -- & 0.04 & -- & -- & -- & 0.11 \\
glass & 0.43 & 0.40 & 0.40 & 0.36 & 0.45 & 0.39 & 0.31 & 0.14 & 0.24 & 0.06 & \textbf{0.51} & 0.36 \\
ionosphere\_UMAP10 & 0.12 & 0.11 & 0.12 & 0.12 & 0.16 & 0.12 & 0.10 & 0.11 & 0.23 & 0.12 & 0.11 & \textbf{0.42} \\
iris & 0.74 & \textbf{0.90} & 0.74 & 0.76 & 0.63 & 0.77 & 0.60 & 0.81 & 0.39 & 0.81 & 0.88 & 0.77 \\
land\_mines & 0.01 & 0.20 & 0.10 & 0.12 & 0.18 & 0.08 & 0.00 & 0.01 & 0.00 & 0.01 & \textbf{0.24} & 0.20 \\
pendigits & 0.69 & 0.68 & 0.69 & 0.69 & 0.68 & 0.73 & 0.00 & 0.74 & 0.38 & 0.71 & \textbf{0.91} & 0.64 \\
seeds & 0.69 & \textbf{0.76} & 0.69 & 0.69 & 0.73 & 0.73 & 0.29 & 0.65 & 0.00 & 0.55 & 0.71 & 0.69 \\
shuttle & 0.00 & 0.00 & 0.26 & 0.16 & -- & -- & 0.00 & 0.08 & 0.16 & 0.12 & \textbf{0.64} & 0.21 \\
wine & 0.43 & \textbf{0.58} & 0.43 & 0.43 & 0.50 & 0.42 & 0.00 & 0.43 & 0.00 & 0.38 & 0.45 & 0.41 \\
yeast & 0.12 & 0.11 & 0.16 & 0.12 & 0.13 & 0.13 & 0.00 & 0.09 & 0.03 & 0.16 & \textbf{0.19} & 0.08 \\
\bottomrule
\end{tabular}

%% file: tables/ARI.tex
\begin{tabular}{lrrrrrrrrrrrr}
\toprule
 & KM & GMM & S-KM & C-KM & COP-KM & Agg & DBSCAN & DEC & DipNSub & DipEnc & LS & Ours \\
dataset &  &  &  &  &  &  &  &  &  &  &  &  \\
\midrule
1d\_gauss & \textbf{1.00} & \textbf{1.00} & \textbf{1.00} & 0.77 & \textbf{1.00} & \textbf{1.00} & \textbf{1.00} & \textbf{1.00} & \textbf{1.00} & \textbf{1.00} & 0.09 & \textbf{1.00} \\
2d\_gauss & 0.76 & \textbf{0.85} & 0.73 & 0.73 & 0.70 & 0.72 & 0.43 & 0.80 & 0.76 & 0.63 & 0.57 & 0.78 \\
6NewsgroupsUMAP10 & 0.61 & 0.60 & 0.63 & 0.60 & 0.58 & 0.55 & 0.12 & 0.60 & 0.52 & 0.60 & 0.51 & \textbf{0.77} \\
MNIST\_UMAP10 & 0.81 & 0.81 & 0.82 & 0.79 & 0.81 & 0.81 & 0.91 & 0.82 & 0.79 & 0.81 & 0.95 & \textbf{0.98} \\
banknote & 0.05 & 0.00 & 0.05 & 0.05 & 0.05 & 0.00 & 0.02 & 0.01 & 0.20 & 0.01 & \textbf{0.60} & 0.02 \\
breast\_cancer & 0.49 & \textbf{0.81} & 0.49 & 0.54 & 0.51 & 0.29 & 0.00 & 0.49 & 0.00 & 0.40 & 0.64 & 0.62 \\
cover\_type & -0.00 & \textbf{0.09} & -0.00 & -- & -- & -- & -- & 0.01 & -- & -- & -- & 0.05 \\
glass & 0.27 & 0.25 & 0.27 & 0.22 & 0.27 & 0.26 & 0.16 & 0.06 & 0.17 & 0.02 & \textbf{0.31} & 0.18 \\
ionosphere\_UMAP10 & 0.15 & 0.14 & 0.15 & 0.15 & 0.20 & 0.15 & -0.07 & 0.13 & 0.17 & 0.15 & 0.11 & \textbf{0.49} \\
iris & 0.72 & \textbf{0.90} & 0.72 & 0.73 & 0.57 & 0.73 & 0.52 & 0.82 & 0.17 & 0.77 & \textbf{0.90} & 0.74 \\
land\_mines & -0.01 & 0.07 & 0.03 & 0.04 & 0.11 & 0.01 & 0.00 & 0.00 & -0.01 & -0.00 & \textbf{0.21} & 0.05 \\
pendigits & 0.58 & 0.54 & 0.60 & 0.60 & 0.56 & 0.55 & 0.00 & 0.60 & 0.21 & 0.56 & \textbf{0.91} & 0.52 \\
seeds & 0.72 & \textbf{0.80} & 0.72 & 0.72 & 0.76 & 0.71 & 0.05 & 0.64 & 0.00 & 0.48 & 0.75 & 0.71 \\
shuttle & 0.00 & 0.00 & 0.46 & 0.10 & -- & -- & 0.00 & 0.04 & 0.00 & 0.09 & \textbf{0.72} & 0.23 \\
wine & 0.37 & \textbf{0.61} & 0.37 & 0.37 & 0.48 & 0.37 & 0.00 & 0.37 & 0.00 & 0.39 & 0.46 & 0.36 \\
yeast & 0.09 & 0.08 & 0.11 & 0.10 & 0.10 & 0.06 & -0.00 & 0.07 & 0.00 & 0.11 & \textbf{0.15} & 0.06 \\
\bottomrule
\end{tabular}

%% file: tables/FMI.tex
\begin{tabular}{lrrrrrrrrrrrr}
\toprule
 & KM & GMM & S-KM & C-KM & COP-KM & Agg & DBSCAN & DEC & DipNSub & DipEnc & LS & Ours \\
dataset &  &  &  &  &  &  &  &  &  &  &  &  \\
\midrule
1d\_gauss & \textbf{1.00} & \textbf{1.00} & \textbf{1.00} & 0.87 & \textbf{1.00} & \textbf{1.00} & \textbf{1.00} & \textbf{1.00} & \textbf{1.00} & \textbf{1.00} & 0.66 & \textbf{1.00} \\
2d\_gauss & 0.80 & \textbf{0.87} & 0.77 & 0.77 & 0.75 & 0.77 & 0.60 & 0.82 & 0.79 & 0.71 & 0.63 & 0.81 \\
6NewsgroupsUMAP10 & 0.68 & 0.68 & 0.70 & 0.67 & 0.66 & 0.64 & 0.44 & 0.68 & 0.60 & 0.68 & 0.59 & \textbf{0.81} \\
MNIST\_UMAP10 & 0.84 & 0.83 & 0.84 & 0.81 & 0.83 & 0.83 & 0.92 & 0.85 & 0.81 & 0.83 & 0.95 & \textbf{0.98} \\
banknote & 0.55 & 0.51 & 0.55 & 0.55 & 0.55 & 0.50 & 0.51 & 0.52 & 0.51 & 0.51 & \textbf{0.81} & 0.60 \\
breast\_cancer & 0.79 & \textbf{0.91} & 0.79 & 0.81 & 0.80 & 0.74 & 0.73 & 0.79 & 0.73 & 0.71 & 0.83 & 0.84 \\
cover\_type & 0.26 & \textbf{0.36} & 0.26 & -- & -- & -- & -- & 0.24 & -- & -- & -- & 0.34 \\
glass & 0.51 & 0.48 & 0.49 & 0.40 & 0.45 & 0.51 & 0.43 & 0.32 & \textbf{0.54} & 0.38 & 0.47 & 0.40 \\
ionosphere\_UMAP10 & 0.59 & 0.58 & 0.59 & 0.59 & 0.61 & 0.59 & 0.54 & 0.58 & 0.48 & 0.59 & 0.57 & \textbf{0.77} \\
iris & 0.81 & \textbf{0.94} & 0.81 & 0.82 & 0.71 & 0.82 & 0.71 & 0.88 & 0.35 & 0.85 & \textbf{0.94} & 0.82 \\
land\_mines & 0.19 & 0.31 & 0.23 & 0.24 & 0.29 & 0.23 & \textbf{0.44} & 0.20 & 0.17 & 0.24 & 0.37 & 0.35 \\
pendigits & 0.62 & 0.59 & 0.64 & 0.64 & 0.61 & 0.61 & 0.32 & 0.65 & 0.28 & 0.61 & \textbf{0.92} & 0.57 \\
seeds & 0.81 & \textbf{0.87} & 0.81 & 0.81 & 0.84 & 0.81 & 0.47 & 0.76 & 0.57 & 0.70 & 0.83 & 0.81 \\
shuttle & 0.80 & 0.81 & 0.84 & 0.61 & -- & -- & 0.80 & 0.80 & 0.04 & 0.55 & \textbf{0.91} & 0.73 \\
wine & 0.58 & \textbf{0.74} & 0.58 & 0.58 & 0.66 & 0.58 & 0.58 & 0.58 & 0.58 & 0.63 & 0.64 & 0.59 \\
yeast & 0.39 & 0.45 & 0.46 & 0.39 & 0.39 & 0.47 & \textbf{0.53} & 0.34 & 0.13 & 0.41 & 0.38 & 0.37 \\
\bottomrule
\end{tabular}

%% file: tables/runtime_s.tex
\begin{tabular}{lrrrrrrrrrrrr}
\toprule
 & KM & GMM & S-KM & C-KM & COP-KM & Agg & DBSCAN & DEC & DipNSub & DipEnc & LS & Ours \\
dataset &  &  &  &  &  &  &  &  &  &  &  &  \\
\midrule
1d\_gauss & 0.05 & 0.04 & \textbf{0.01} & 1.50 & 3.59 & 4.81 & 0.22 & 17.92 & 0.02 & 210.90 & 0.03 & 0.12 \\
2d\_gauss & 0.02 & 0.14 & \textbf{0.01} & 5.89 & 4.80 & 1.40 & 0.07 & 12.17 & 0.02 & 76.04 & 0.03 & 0.11 \\
6NewsgroupsUMAP10 & 0.02 & 0.11 & \textbf{0.01} & 2.25 & 1.52 & 0.42 & 0.09 & 6.76 & 0.38 & 44.44 & 0.03 & 0.12 \\
MNIST\_UMAP10 & 0.01 & 0.03 & \textbf{0.00} & 0.48 & 0.47 & 0.03 & 0.01 & 2.58 & 0.57 & 14.27 & 0.02 & 2.60 \\
banknote & 0.01 & 0.01 & \textbf{0.00} & 0.20 & 0.08 & 0.02 & \textbf{0.00} & 2.04 & 0.16 & 19.79 & 0.02 & 0.02 \\
breast\_cancer & 0.01 & 0.10 & \textbf{0.00} & 0.05 & 0.12 & \textbf{0.00} & 0.01 & 0.85 & 0.19 & 8.88 & \textbf{0.00} & 0.01 \\
cover\_type & 1.04 & 81.25 & \textbf{0.99} & -- & -- & -- & -- & 570.80 & -- & -- & -- & 26.27 \\
glass & \textbf{0.00} & 0.01 & \textbf{0.00} & 0.07 & 0.64 & \textbf{0.00} & \textbf{0.00} & 0.37 & 0.04 & 2.26 & 0.01 & 0.79 \\
ionosphere\_UMAP10 & \textbf{0.00} & \textbf{0.00} & \textbf{0.00} & 0.02 & 0.02 & \textbf{0.00} & \textbf{0.00} & 0.59 & 0.13 & 5.92 & 0.01 & 0.01 \\
iris & 0.01 & 0.01 & \textbf{0.00} & 0.03 & 0.01 & \textbf{0.00} & \textbf{0.00} & 0.32 & 0.07 & 2.41 & 0.02 & \textbf{0.00} \\
land\_mines & 0.01 & 0.03 & \textbf{0.00} & 0.11 & 0.17 & \textbf{0.00} & \textbf{0.00} & 0.58 & 0.06 & 3.48 & 0.02 & 0.70 \\
pendigits & 0.04 & 2.00 & \textbf{0.01} & 22.32 & 13.04 & 2.10 & 0.43 & 14.04 & 22.51 & 88.87 & 2.16 & 0.14 \\
seeds & \textbf{0.00} & 0.01 & \textbf{0.00} & 0.03 & 0.02 & \textbf{0.00} & \textbf{0.00} & 0.35 & 0.02 & 2.88 & 0.01 & \textbf{0.00} \\
shuttle & 0.17 & 0.52 & \textbf{0.03} & 8.16 & -- & -- & 0.55 & 46.73 & 203.89 & 521.01 & 0.54 & 0.77 \\
wine & \textbf{0.00} & 0.01 & \textbf{0.00} & 0.04 & 0.04 & \textbf{0.00} & \textbf{0.00} & 0.32 & 0.03 & 3.04 & 0.02 & \textbf{0.00} \\
yeast & 0.02 & 0.13 & \textbf{0.01} & 1.62 & 0.24 & 0.02 & 0.04 & 2.22 & 0.30 & 15.46 & 0.02 & 0.02 \\
\bottomrule
\end{tabular}

%% file: tables/full_eval_proposed.tex
\begin{tabular}{lrrrrrlr}
\toprule
Dataset & ARI & NMI & V-Measure & FMI & Purity & Rejected ($n/N$) & Rejection Rate \\
\midrule
1d\_gauss & 0.91 & 0.87 & 0.87 & 0.95 & 0.98 & 955/17528 & 0.05 \\
2d\_gauss & 0.80 & 0.83 & 0.83 & 0.82 & 0.89 & 644/10300 & 0.06 \\
6NewsgroupsUMAP10 & 0.54 & 0.60 & 0.60 & 0.62 & 0.77 & 1078/5881 & 0.18 \\
MNIST\_UMAP10 & 0.64 & 0.80 & 0.80 & 0.68 & 0.84 & 394/1797 & 0.22 \\
banknote & 0.05 & 0.08 & 0.08 & 0.57 & 0.64 & 99/1372 & 0.07 \\
breast\_cancer & 0.60 & 0.46 & 0.46 & 0.81 & 0.90 & 39/569 & 0.07 \\
cover\_type & 0.05 & 0.11 & 0.11 & 0.34 & 0.49 & 7407/581012 & 0.01 \\
glass & 0.14 & 0.32 & 0.32 & 0.34 & 0.53 & 19/214 & 0.09 \\
ionosphere\_UMAP10 & 0.26 & 0.24 & 0.24 & 0.60 & 0.82 & 93/351 & 0.27 \\
iris & 0.42 & 0.51 & 0.51 & 0.59 & 0.75 & 41/150 & 0.27 \\
land\_mines & 0.05 & 0.20 & 0.20 & 0.34 & 0.42 & 0/338 & 0.00 \\
pendigits & 0.51 & 0.64 & 0.64 & 0.56 & 0.69 & 0/10992 & 0.00 \\
seeds & 0.68 & 0.63 & 0.63 & 0.78 & 0.89 & 13/210 & 0.06 \\
shuttle & 0.34 & 0.31 & 0.31 & 0.73 & 0.83 & 1647/43500 & 0.04 \\
wine & 0.35 & 0.39 & 0.39 & 0.58 & 0.72 & 8/178 & 0.04 \\
yeast & 0.10 & 0.14 & 0.14 & 0.42 & 0.42 & 52/1484 & 0.04 \\
\bottomrule
\end{tabular}

%% file: tables/full_eval_compare.tex
\begin{tabular}{llrrrrrrrrrrrr}
\toprule
Metric & Dataset & KM & GMM & S-KM & C-KM & COP-KM & Agg & DBSCAN & DEC & DipNSub & DipEnc & LS & Ours \\
\midrule
\multirow[t]{16}{*}{ARI} & 1d\_gauss & \textbf{1.00} & \textbf{1.00} & \textbf{1.00} & 0.77 & \textbf{1.00} & \textbf{1.00} & \textbf{1.00} & \textbf{1.00} & \textbf{1.00} & \textbf{1.00} & 0.09 & 0.91 \\
 & 2d\_gauss & 0.76 & \textbf{0.85} & 0.73 & 0.73 & 0.70 & 0.72 & 0.52 & 0.80 & 0.74 & 0.61 & 0.54 & 0.80 \\
 & 6NewsgroupsUMAP10 & 0.61 & 0.60 & \textbf{0.63} & 0.60 & 0.58 & 0.55 & 0.12 & 0.60 & 0.52 & 0.60 & 0.51 & 0.54 \\
 & MNIST\_UMAP10 & 0.81 & 0.81 & 0.82 & 0.79 & 0.81 & 0.81 & 0.91 & 0.82 & 0.79 & 0.81 & \textbf{0.95} & 0.64 \\
 & banknote & 0.05 & 0.00 & 0.05 & 0.05 & 0.05 & 0.00 & 0.01 & 0.01 & 0.20 & 0.01 & \textbf{0.60} & 0.05 \\
 & breast\_cancer & 0.49 & \textbf{0.81} & 0.49 & 0.54 & 0.51 & 0.29 & 0.00 & 0.49 & 0.00 & 0.40 & 0.64 & 0.60 \\
 & cover\_type & -0.00 & \textbf{0.09} & -0.00 & -- & -- & -- & -- & 0.01 & -- & -- & -- & 0.05 \\
 & glass & 0.27 & 0.25 & 0.27 & 0.22 & 0.27 & 0.26 & 0.16 & 0.06 & 0.17 & 0.02 & \textbf{0.31} & 0.14 \\
 & ionosphere\_UMAP10 & 0.15 & 0.14 & 0.15 & 0.15 & 0.20 & 0.15 & -0.06 & 0.13 & 0.17 & 0.15 & 0.11 & \textbf{0.26} \\
 & iris & 0.72 & \textbf{0.90} & 0.72 & 0.73 & 0.57 & 0.73 & 0.52 & 0.82 & 0.17 & 0.77 & \textbf{0.90} & 0.42 \\
 & land\_mines & -0.01 & 0.07 & 0.03 & 0.04 & 0.11 & 0.01 & 0.00 & 0.00 & -0.01 & -0.00 & \textbf{0.21} & 0.05 \\
 & pendigits & 0.58 & 0.54 & 0.60 & 0.60 & 0.56 & 0.55 & 0.00 & 0.60 & 0.21 & 0.56 & \textbf{0.91} & 0.51 \\
 & seeds & 0.72 & \textbf{0.80} & 0.72 & 0.72 & 0.76 & 0.71 & 0.06 & 0.64 & 0.00 & 0.48 & 0.75 & 0.68 \\
 & shuttle & 0.00 & 0.00 & 0.46 & 0.10 & -- & -- & 0.00 & 0.04 & 0.00 & 0.09 & \textbf{0.71} & 0.34 \\
 & wine & 0.37 & \textbf{0.61} & 0.37 & 0.37 & 0.48 & 0.37 & 0.00 & 0.37 & 0.00 & 0.39 & 0.46 & 0.35 \\
 & yeast & 0.09 & 0.08 & 0.11 & 0.10 & 0.10 & 0.06 & 0.01 & 0.07 & 0.00 & 0.12 & \textbf{0.13} & 0.10 \\
\cline{1-14}
\multirow[t]{16}{*}{NMI} & 1d\_gauss & \textbf{1.00} & \textbf{1.00} & \textbf{1.00} & 0.80 & \textbf{1.00} & \textbf{1.00} & 0.99 & \textbf{1.00} & 0.99 & 0.99 & 0.16 & 0.87 \\
 & 2d\_gauss & 0.87 & \textbf{0.92} & 0.83 & 0.83 & 0.82 & 0.85 & 0.74 & 0.85 & 0.83 & 0.78 & 0.68 & 0.83 \\
 & 6NewsgroupsUMAP10 & 0.61 & 0.61 & \textbf{0.62} & 0.60 & 0.60 & 0.61 & 0.28 & \textbf{0.62} & 0.57 & 0.61 & 0.51 & 0.60 \\
 & MNIST\_UMAP10 & 0.90 & 0.90 & 0.90 & 0.87 & 0.89 & 0.90 & 0.93 & 0.90 & 0.87 & 0.90 & \textbf{0.95} & 0.80 \\
 & banknote & 0.03 & 0.01 & 0.03 & 0.03 & 0.03 & 0.00 & 0.17 & 0.01 & 0.33 & 0.01 & \textbf{0.57} & 0.08 \\
 & breast\_cancer & 0.46 & \textbf{0.71} & 0.46 & 0.50 & 0.48 & 0.32 & 0.00 & 0.46 & 0.00 & 0.41 & 0.52 & 0.46 \\
 & cover\_type & 0.07 & \textbf{0.17} & 0.07 & -- & -- & -- & -- & 0.04 & -- & -- & -- & 0.11 \\
 & glass & 0.43 & 0.40 & 0.40 & 0.36 & 0.45 & 0.39 & 0.31 & 0.14 & 0.24 & 0.06 & \textbf{0.51} & 0.32 \\
 & ionosphere\_UMAP10 & 0.12 & 0.11 & 0.12 & 0.12 & 0.16 & 0.12 & 0.09 & 0.11 & 0.23 & 0.12 & 0.11 & \textbf{0.24} \\
 & iris & 0.74 & \textbf{0.90} & 0.74 & 0.76 & 0.63 & 0.77 & 0.60 & 0.81 & 0.39 & 0.81 & 0.88 & 0.51 \\
 & land\_mines & 0.01 & 0.20 & 0.10 & 0.12 & 0.18 & 0.08 & 0.00 & 0.01 & 0.00 & 0.01 & \textbf{0.24} & 0.20 \\
 & pendigits & 0.69 & 0.68 & 0.69 & 0.69 & 0.68 & 0.73 & 0.00 & 0.74 & 0.38 & 0.71 & \textbf{0.91} & 0.64 \\
 & seeds & 0.69 & \textbf{0.76} & 0.69 & 0.69 & 0.73 & 0.73 & 0.30 & 0.65 & 0.00 & 0.55 & 0.71 & 0.63 \\
 & shuttle & 0.00 & 0.00 & 0.26 & 0.16 & -- & -- & 0.00 & 0.08 & 0.16 & 0.12 & \textbf{0.62} & 0.31 \\
 & wine & 0.43 & \textbf{0.58} & 0.43 & 0.43 & 0.50 & 0.42 & 0.00 & 0.43 & 0.00 & 0.38 & 0.45 & 0.39 \\
 & yeast & 0.12 & 0.11 & \textbf{0.16} & 0.12 & 0.13 & 0.13 & 0.02 & 0.09 & 0.04 & \textbf{0.16} & \textbf{0.16} & 0.14 \\
\bottomrule
\end{tabular}